\documentclass[11pt]{article}

\usepackage[final]{acl}

\usepackage{times}
\usepackage{latexsym}

\usepackage[T1]{fontenc}
\usepackage{enumitem}
\usepackage{xspace}
\usepackage{amssymb}

\usepackage[utf8]{inputenc}
\usepackage{amsmath}
\usepackage{microtype}

\usepackage{inconsolata}

\usepackage{graphicx}

\usepackage{microtype}
\usepackage{amsmath}
\usepackage{amssymb}
\usepackage{amsfonts}
\usepackage{amsthm}

\usepackage{xcolor}
\usepackage{graphicx}
\usepackage{float}
\usepackage{wrapfig}
\usepackage{subcaption}
\usepackage{colortbl}

\usepackage{booktabs}
\usepackage{array}
\usepackage{tabularx}
\usepackage{threeparttable}
\usepackage{multirow}
\usepackage{multicol}
\usepackage{makecell}
\usepackage{longtable}
\usepackage{ragged2e}
\usepackage{adjustbox}
\usepackage{caption}
\usepackage{calc}

\usepackage{enumitem}
\usepackage{nicefrac}

\usepackage{algorithm}

\makeatletter
\@ifpackageloaded{algorithmic}{
  \let\algorithm ic\relax
  
  \let\algorithmicrequire\relax
  \let\algorithmicensure\relax
}{}
\makeatother

\usepackage{algpseudocode}
\algrenewcommand\algorithmicrequire{\textbf{Input:}}
\algrenewcommand\algorithmicensure{\textbf{Output:}}

\usepackage[most]{tcolorbox}
\usepackage{etoolbox}

\usepackage{listings}
\newtcblisting{breakablelisting}{
  listing only,
  breakable,
  colback=white,
  colframe=black,
  boxrule=0.4pt,
  arc=0pt,
  left=2pt,
  right=2pt,
  top=2pt,
  bottom=2pt,
  listing options={
    basicstyle=\ttfamily\small,
    columns=fullflexible,
    keepspaces=true,
    showstringspaces=false,
    breaklines=true,
    breakindent=0pt,
    breakautoindent=false,
    prebreak=\mbox{},
    postbreak=\mbox{},
  }
}

\newtcolorbox{exbox}[1]{
  breakable,
  title=#1,
  colback=white,
  colframe=green!60!black,
  coltitle=white,
  colbacktitle=green!60!black,
  fonttitle=\bfseries
}

\definecolor{prompttitlegray}{gray}{0.25}

\newtcblisting{promptbox}{
  breakable,
  colback=white,
  colframe=black,
  boxrule=0.5pt,
  sharp corners,
  left=4pt,right=4pt,top=4pt,bottom=4pt,
  listing only,
  listing options={
    basicstyle=\ttfamily\small,
    columns=fullflexible,
    keepspaces=true,
    breaklines=true,
  }
}

\usepackage{pgfplotstable}

\definecolor{lightblue1}{rgb}{0.97, 0.985, 1} 
\definecolor{lightblue2}{rgb}{0.92, 0.965, 1} 
\definecolor{lightblue3}{rgb}{0.84, 0.93, 1}
\definecolor{lightblue4}{rgb}{0.74, 0.87, 1}
\definecolor{lightblue5}{rgb}{0.64, 0.81, 1}
\definecolor{lightblue6}{rgb}{0.54, 0.75, 1}

\definecolor{lightgreen1}{rgb}{0.97, 1.00, 0.97}
\definecolor{lightgreen2}{rgb}{0.92, 0.98, 0.92}
\definecolor{lightgreen3}{rgb}{0.84, 0.95, 0.84}
\definecolor{lightgreen4}{rgb}{0.74, 0.91, 0.74}
\definecolor{lightgreen5}{rgb}{0.64, 0.86, 0.64}
\definecolor{lightgreen6}{rgb}{0.54, 0.81, 0.54}

\definecolor{lightorange1}{rgb}{1.00, 0.98, 0.95}
\definecolor{lightorange2}{rgb}{1.00, 0.95, 0.85}
\definecolor{lightorange3}{rgb}{1.00, 0.90, 0.70}
\definecolor{lightorange4}{rgb}{1.00, 0.85, 0.55}
\definecolor{lightorange5}{rgb}{1.00, 0.80, 0.40}
\definecolor{lightorange6}{rgb}{1.00, 0.75, 0.30}

\definecolor{lightpurple1}{rgb}{0.985, 0.97, 1.00}
\definecolor{lightpurple2}{rgb}{0.96, 0.92, 1.00}
\definecolor{lightpurple3}{rgb}{0.93, 0.84, 1.00}
\definecolor{lightpurple4}{rgb}{0.87, 0.74, 1.00}
\definecolor{lightpurple5}{rgb}{0.81, 0.64, 1.00}
\definecolor{lightpurple6}{rgb}{0.75, 0.54, 1.00}

\definecolor{lightred1}{rgb}{1.00, 0.97, 0.97}
\definecolor{lightred2}{rgb}{1.00, 0.92, 0.92}
\definecolor{lightred3}{rgb}{1.00, 0.84, 0.84}
\definecolor{lightred4}{rgb}{1.00, 0.74, 0.74}
\definecolor{lightred5}{rgb}{1.00, 0.64, 0.64}
\definecolor{lightred6}{rgb}{1.00, 0.54, 0.54}

\definecolor{lightcyan1}{rgb}{0.97, 1.00, 1.00}
\definecolor{lightcyan2}{rgb}{0.92, 0.98, 0.98}
\definecolor{lightcyan3}{rgb}{0.84, 0.95, 0.96}
\definecolor{lightcyan4}{rgb}{0.74, 0.91, 0.94}
\definecolor{lightcyan5}{rgb}{0.64, 0.87, 0.92}
\definecolor{lightcyan6}{rgb}{0.54, 0.83, 0.90}

\usepackage{tikz}
\usetikzlibrary{tikzmark}
\definecolor{deepblue}{RGB}{48, 58, 82}
\makeatletter
\newcommand*\myfontsize{
  \@setfontsize\myfontsize{10}{8}
}
\makeatother
\newcommand{\mytextbox}[2]{\tikzmarknode[draw=#1,thick,inner sep=2pt]{test}{\myfontsize #2}}
\newcommand{\true}{\mytextbox{deepblue}{\textbf{\textcolor{deepblue}{True}}}}
\newcommand{\false}{\mytextbox{deepblue}{\textbf{\textcolor{deepblue}{False}}}}
\newcommand{\answerl}{\mytextbox{deepblue}{\textbf{\textcolor{deepblue}{<answer>}}}}
\newcommand{\answerr}{\mytextbox{deepblue}{\textbf{\textcolor{deepblue}{</answer>}}}}

\newcommand{\actionl}{\mytextbox{deepblue}{\textbf{\textcolor{deepblue}{<python>}}}}
\newcommand{\actionr}{\mytextbox{deepblue}{\textbf{\textcolor{deepblue}{</python>}}}}

\title{AgentV-RL: Scaling Reward Modeling with Agentic Verifier}

\author{\textbf{Jiazheng Zhang}$^{1}\thanks{{ }{ }Equal contribution.} \: \:  \:$,
\textbf{Ziche Fu}$^{1 *}$,
\textbf{Zhiheng Xi}$^{1 *}$,
\textbf{Wenqing Jing}$^{1}$, 
\ \textbf{Mingxu Chai}$^{1}$,
\textbf{Wei He}$^{1}$,
\\
\textbf{Guoqiang Zhang}$^{1}$,
\ \textbf{Chenghao Fan}$^{2}$,
\ \textbf{Chenxin An}$^{3}$,
\textbf{Wenxiang Chen}$^{1}$,
\ \textbf{Zhicheng Liu}$^{4}$,
\\
\ \textbf{Haojie Pan}$^{4}$,
\ \textbf{Dingwei Zhu}$^{1}$,
\ \textbf{Tao Gui}$^{5,6}$\thanks{{ }{ }Corresponding author.},
\ \textbf{Qi Zhang}$^{5,6}$,
\ \textbf{Xuanjing Huang}$^{5,6}$
\\
$^{1}$College of Computer Science and Artificial Intelligence, Fudan University \\ $^{2}$ Huazhong University of Science and Technology  $^{3}$ The University of Hong Kong \\ $^{4}$ ByteDance Seed   \ 
$^{5}$ Institute of Trustworthy Embodied AI, Fudan University \\
$^6$ Shanghai Key Laboratory of Multimodal Embodied AI \\
\texttt{jzzhang24@m.fudan.edu.cn},  \texttt{tgui@fudan.edu.cn}
}

\begin{document}
\maketitle
\begin{abstract}

Verifiers have been demonstrated to enhance LLM reasoning via test-time scaling (TTS).
Yet, they face significant challenges in complex domains. Error propagation from incorrect intermediate reasoning can lead to false positives for seemingly plausible solutions, while lacking external grounding makes verifiers unreliable on computation or knowledge-intensive tasks.
To address these challenges, we propose \textbf{Agentic Verifier}, a framework that transforms reward modeling into a multi-turn, tool-augmented deliberative process.
We introduce complementary forward and backward agents: one traces solutions from premises to conclusions, while the other re-checks conclusions against their underlying premises.
This bidirectional process enables a comprehensive, reliable, and interpretable assessment of solutions.
To facilitate practical deployment, we propose \textbf{AgentV-RL}.
Through proactive exploration and reinforcement learning, the verifier autonomously interleaves tool-use with internal reasoning.
Extensive experiments show that Agentic Verifier yields consistent performance gains under both parallel and sequential TTS.
Notably, our 4B variant surpasses state-of-the-art ORMs by 25.2\%, positioning it as a promising paradigm for agentic reward modeling.
Our code is available at \href{https://github.com/JiazhengZhang/AgentV-RL}{GitHub}.

\end{abstract}

\section{Introduction}
\label{sec:intro}

\begin{figure}[t]
\centering
\includegraphics[width=0.95\columnwidth]{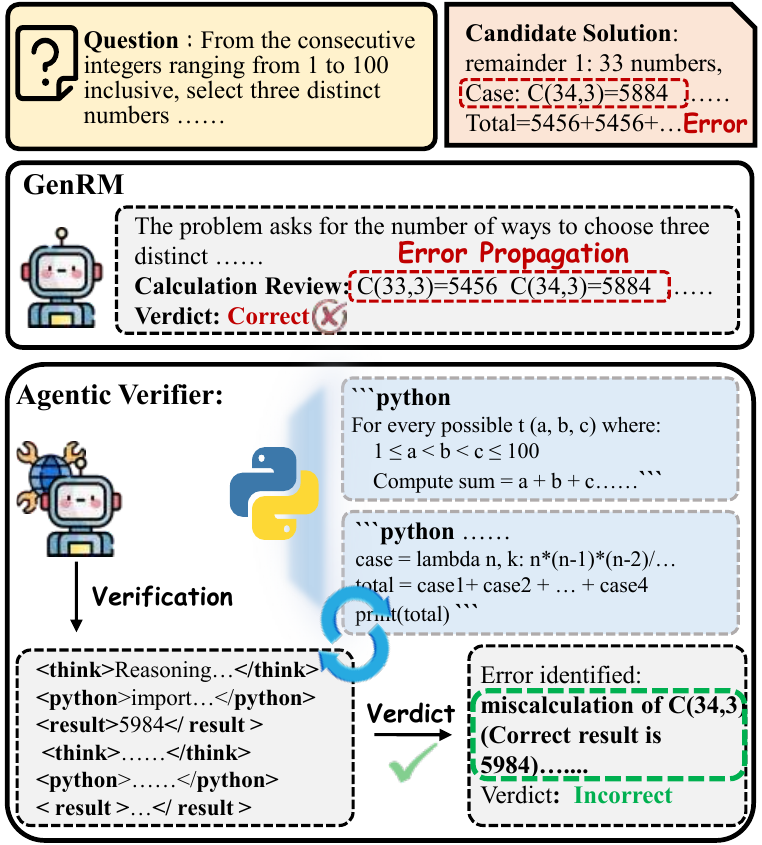}
\caption{
Agentic Verifier v.s. GenRM: while GenRM suffers from error propagation and is misled by incorrect solutions, Agentic Verifier ensures a rigorous review with external grounding.
}
\label{fig:Intro}
\vspace{-6mm}
\end{figure}

Recent milestones on the International Mathematics Olympiad by OpenAI and Google~\citep{OAI:2025:TwitterIMO, Google:2025:GoogleIMO} highlight the rapid ascent of reasoning models like Gemini-3 and DeepSeek-Math-V2~\citep{Shao:2025:deepseekmathv2}. 
To push the capability boundaries of LLM, scaling inference-time compute has emerged as a prevalent trend~\citep{Niklas:2025:s1, huang:2025:GeminiPipline, chen:2025:SeedProver}. Whether through parallel methods (e.g., Best-of-$N$) or sequential refinement, the efficacy of Test-Time Scaling (TTS) is fundamentally dependent on the Reward Model, i.e., verifier, serving as the critical compass for guiding the search process and discerning solution quality.

\par
Existing reward models, represented by outcome-level RM (ORM, \citealp{zhong2025comprehensivesurveyrewardmodels,wang2024secretsrlhflargelanguage,Zhang:2025:CRM}) and process-level RM (PRM, \citealp{zhang2025lessonsdevelopingprocessreward,chen2025better,Lightman:2024:MATH500, cui2025processreinforcementimplicitrewards}), only output scalar values while lacking interpretability. 
Though recent efforts utilize the next-token prediction objective for training Generative Reward Model~(GenRM,~\citealp{Mahan:2025:GRM,Zhang:2025:GenerativeVerifiers,Chen:2025:RM-R1,Liu:2025:DeepseekGRM}), this line of work typically employs single-turn reasoning to assess the candidate solution. Specifically, this prevalent paradigm suffers from critical limitations in Figure~\ref{fig:Intro}: 
(1)~\textbf{Error propagation}, since LLM are mostly trained on correct or near-correct solutions, they struggle to obtain correct verdict conditioned on flawed solution, easily misled by superficially plausible but incorrect answers~\citep{Zhang:2025:GenerativeVerifiers}. (2)~\textbf{External grounding}, verifiers often struggle with computation-intensive or knowledge-heavy domains. Without integration with symbolic solvers or external tools~\citep{Dong:2025:AEPO,Feng:2025:ReTool,Dong:2025:ToolStar}, they are prone to hallucinations that render unstable performance.
These limitations highlight the trend for a paradigm shift towards agentic reward modeling through multi-turn reasoning integrated with external tools.

\par
To bridge this gap, we propose \textbf{Agentic Verifier}, a multi-agent framework that emulates rigorous human-like checking. Inspired by mathematical proof strategies, Agentic Verifier coordinates two specialized agents: the forward agent and the backward agent.
The forward agent, responsible for sufficiency checking by tracing the logical flow from premises to conclusion, and the backward agent performs necessity checking by validating that the whether the solution is grounded in the problem constraints.
Both agents are empowered with multi-turn reasoning and tool-augmented verification capabilities: they can iteratively decompose complex solutions into verifiable sub-steps, invoke external tools such as code interpreters for numerical calculation. 
Together, this collaborative mechanism allows a comprehensive review that ensures a reliable verdict, proactively identifying immediate flaws or unwarranted content.

\par
To address the challenges inherent in training multi-agent systems, we introduce \textbf{AgentV-RL} to distill this multi-agent capacity into a single model.
This recipe comprises an end-to-end synthetic data engine that automatically generates verification trajectories and conducts quality control, and a two-stage training schema to unlock the reasoning potential.
By automatically constructing verification trajectories that encompass a broad spectrum of logical and computational challenges, this synthetic engine not only alleviates data scarcity but also ensures comprehensive coverage of difficult reasoning patterns.
Building on this foundation, our two-stage training recipe is designed to empower the verifier multi-turn, long-horizon, and tool-integrated reasoning by rejection sampling fine-tuning followed by RL.

\par
Finally, we conduct thorough experiments to review the effectiveness of Agentic Verifier under extensive settings. 
For parallel TTS, i.e., BoN (Best-of-N), our Agentic Verifier outperforms proprietary reasoning models and off-the-shelf BT-RMs on mathematics.  
Remarkably, our 4B variant consistently outperforms INF-ORM-Llama3.1-70B~\citep{INF_ORM_Llama3.1_70B}, an outcome reward model with ten times more parameters.
For sequential TTS, Agentic Verifier serves as an effective critique model and provides desirable feedback to help correct the actor's flaws. In-depth analysis in Sec.~\ref{sec:exp} confirms its efficacy and improves actor performance. Overall, our main contributions are:
\begin{itemize}[leftmargin=*]
\item We offer an agentic paradigm for reward modeling, \textbf{Agentic Verifier}, orchestrating two specialized agents to proactively identify flaws in seemingly plausible solutions.
\item We introduce \textbf{AgentV-RL}, a scalable recipe that distills the capabilities of multi-agents into a single LLM, empowering the verifier with multi-turn, long-horizon, and tool-integrated reasoning.
\item Empirical experiments demonstrate the efficacy of the proposed method. Notably, our 4B variant outperforms the state-of-the-art ORM and attains up to 25.2\% improvements.
\end{itemize}

\begin{figure*}[t]
\centering
\includegraphics[width=\linewidth]{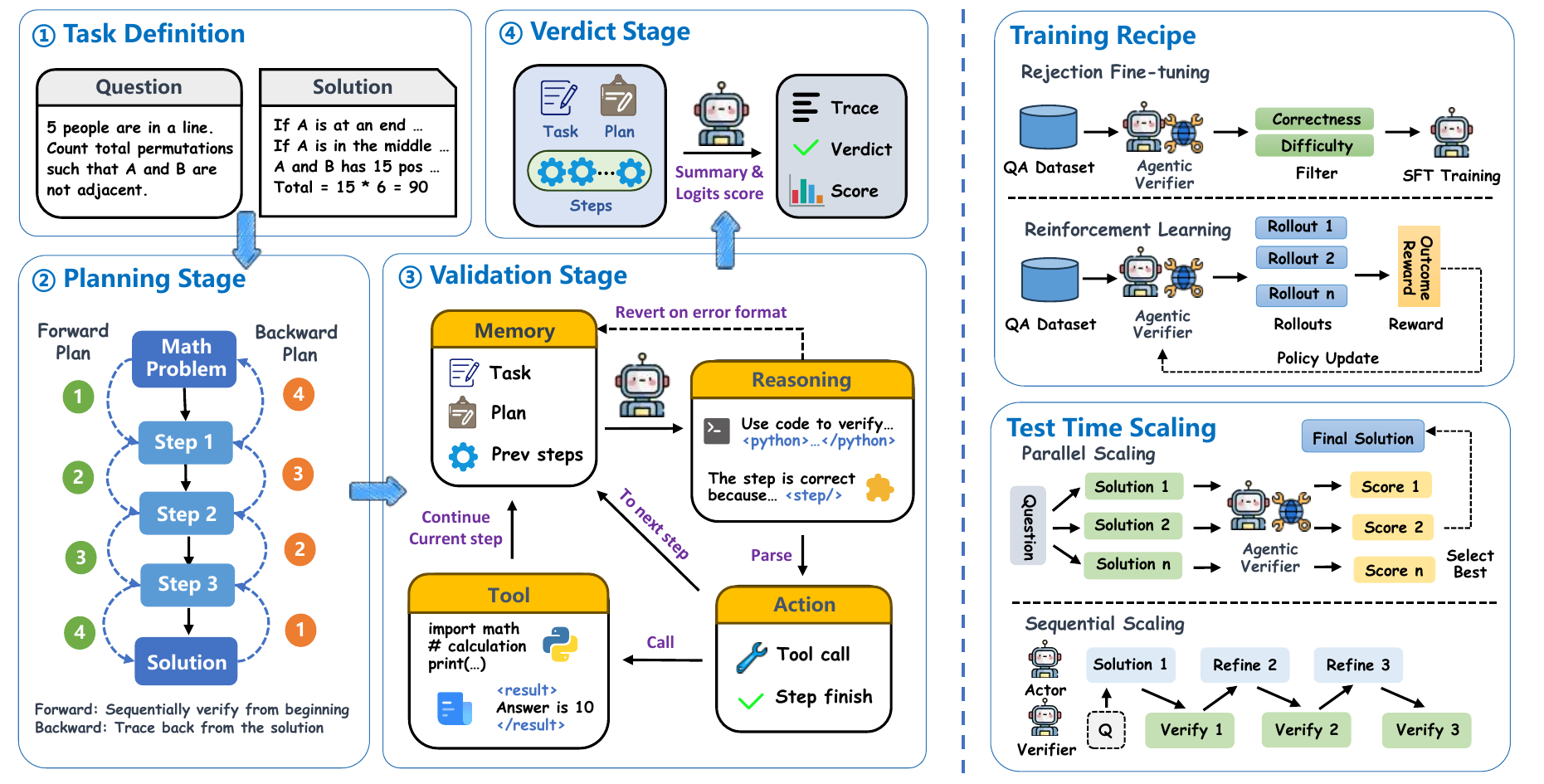} 
\caption {
\textbf{Overview of Agentic Verifier's architecture.}
Agentic Verifier coordinates forward and backward agents with multi-turn reasoning and tool-augmented verification for reliable validation.
}
\label{fig:Agentic_Verifier}
\vspace{-4mm}
\end{figure*}

\section{Related Work}

\paragraph{Reward Model (RM).}
RMs play a pivotal role in aligning large language models (LLMs) with human preferences. 
Traditional outcome-level RMs, assign scalar rewards to the complete response based on preference rankings~\citep{zhong2025comprehensivesurveyrewardmodels,wang2024secretsrlhflargelanguage}.
To address the limitations of sparse supervision, PRMs provide dense signals by supervising intermediate steps~\citep{zhang2025lessonsdevelopingprocessreward,chen2025better,Lightman:2024:MATH500,cui2025processreinforcementimplicitrewards}.
Recent work has explored generative reward models (GenRMs), which reformulate reward modeling as next-token prediction that generates natural-language feedback~\citep{mahan2024GRM,chen2025judgelrm,zhang2024generative-verifiers,li2025genRMs}.
Building on this paradigm, rubric-based GenRMs  dynamically construct task-specific rubrics and reason about evaluation criteria~\citep{guo2025critiq,liu2025inference-time-scaling, Chen:2025:RM-R1,guo2025reward-reasoning-model}.
In parallel, several approaches augment RMs with tool under LLM-as-Judge framework~\citep{Li:2024:ToolRM,Xu:2025:LLMJudgeTIR,peng2025agenticrewardmodeling}.
However, existing methods either do not tightly integrate tool execution into the reasoning process or fail to provide point-wise feedback required for test-time scaling (TTS).
In contrast, our work reformulates verification as an agentic, multi-turn process, enabling test-time exploration and reliable assessment.

\paragraph{Test-Time Scaling and Verifiers.}
Recent work has shown that scaling inference-time compute can substantially improve LLM reasoning, and test-time scaling (TTS) has emerged as a general paradigm for both parallel selection and sequential refinement \cite{muennighoff2025s1}. Within this setting, critique-based methods use auxiliary models to guide actor correction and self-improvement at test time \cite{xi2024enhancing}, while more recent studies show that reward models and process verifiers themselves can also benefit from additional inference-time computation \cite{liu2025inference,khalifa2025process}. Our work is closely related to this line, but differs in that we cast verification as a bidirectional, multi-turn, tool-augmented process, enabling both sufficiency and necessity checking under parallel and sequential TTS.

\section{Method}
\label{sec:method}
In this paper, we focus on the interactive TTS behavior between the actor and verifier on mathematics. The actor is involved in solving problems, while the verifier provides supervisory feedback on the generated solution chains. 

\paragraph{Parallel Scaling.} Best-of-N (BoN) has emerged as a prevalent parallel sampling strategy that leverages a verifier to select high-quality solutions \citep{Gui:2024:BoNAlignment,Kang:2025:ScalableBoN}. Specifically, for a given input $x$, the actor samples $k$ candidate solutions, denoted as $\{y^{(j)}\}_{j=1}^{k}$. Subsequently, the verifier $\pi_{\psi}$ assesses these candidates and generates a verifying rationale $f$ to evaluate their correctness. The solution with the highest confidence score in $\pi_{\psi}$ is then selected. This confidence score is defined as the likelihood of the  \true\ token  within the verifying critique $f$, calculated as:
\begin{align}
l(x, y^{(j)})&=\pi_{\psi}(\text{\true} \mid x, y^{(j)}, f^{(j)},\mathbf{I}) \nonumber \\
f^{(j)}\sim &\pi_{\psi}(x,y^{(j)}, \mathbf{I}),
\label{eq:boN}
\end{align}
where $\mathbf{I}$ is the instruction: Is the solving process correct? (\raisebox{0.2ex}{\true} / \raisebox{0.2ex}{\false}). 

\paragraph{Sequential Scaling.} Given query $x$ and an initial solution $y_{0}$, the verifier analyzes the solving process and provides supervisory critique, i.e., $f_{0} \sim \pi_{\psi}(x, y_{0})$. The actor then receives this feedback and refines its solution as $y_{1}$. This refinement cycle can be repeated for multiple rounds until a correct answer is obtained or stopping criterion, formally expressed as 
\begin{equation}
y_{t} \sim \pi_{\theta}(x, y_{0}, f_{0},\ldots,y_{t-1},f_{t-1}).
\label{eq:refine}
\end{equation}

\subsection{Agentic Verifier}
Prevalent GenRMs suffer from error propagation and attention drift due to the single-turn reasoning shown in Figure~\ref{fig:Intro}, which limits their efficacy for reliable verification.
As depicted in Figure~\ref{fig:Agentic_Verifier}, we propose \textbf{Agentic Verifier} coordinating complementary forward and backward agents that trace solutions from premises to conclusions and re-examine conclusions against their premises.
Both agents are accommodated with multi-turn reasoning and tool-augmented verification: they can iteratively decompose complex solutions into verifiable sub-steps, and invoke external tools like code for numerical calculation. 
We provide comprehensive details in Appendix~\ref{sec:appendix_ours_prompts}.

\begin{figure*}[t]
\centering
\includegraphics[width=\linewidth]{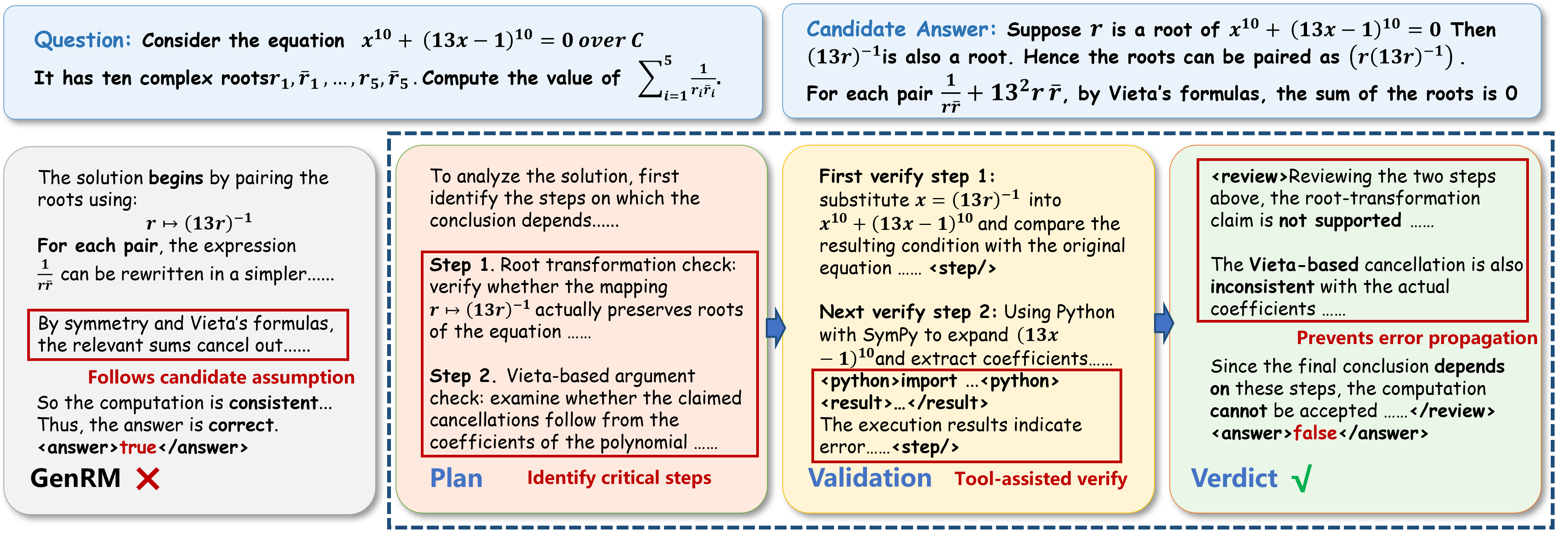}
\caption{
\textbf{Case study comparing GenRM and Agentic Verifier.}
The example highlights the error propagation of GenRM, while our method obtains the correct verdict. More examples are provided in Appendix~\ref{sec:appendix_case}.}
\label{fig:case_study}
\vspace{-5mm}
\end{figure*}
\subsection{Task Definition}
Formally, given $\mathcal{D}=\left\{\left(x, g\right)\right\}$, where $x$ and $g$ is the question and ground truth. The actor $\pi_{\theta}$ generates the solution  $y \sim \pi_{\theta}(\cdot| x)$ autoregressively, which consists of the step-by-step CoT rationale and the final answer $\tilde{y}$. 
Following~\citet{Xi:20254:Critique, Zhang:2025:GenerativeVerifiers}, we investigate two popular Test-Time Scaling paradigms: Parallel Scaling and Sequential Scaling.

\paragraph{Forward Agent.}
Starting from the problem premises, forward agent sequentially traces the solution path to review whether each step in the solution is correct, and validate whether the preceding step constitutes a sufficient condition for the subsequent derivation.
To inspire the agent's capabilities, we adopt a "Plan-Validate-Verdict" strategy in Figure~\ref{fig:case_study}:
\begin{itemize}[leftmargin=*]
\item \textbf{Planning.} This phase aims at decomposing the overcomplicated reasoning solution into a sequence of atomic, verifiable sub-steps $\Pi = \{v_1, v_2, \dots, v_n\}$. The agent formulates an explicit plan specifying the logical flow for the subsequent stage. This plan includes specific checkpoints and necessary calculations, establishing a structured template for the whole verification.

\item \textbf{Validation.} After receiving the plan $\Pi$, the agent examines the correctness of each atomic step $v_i$, ensuring the logical sufficiency between $v_{i-1}$ and $v_{i}$.
The agent is empowered to invoke Python interpreter and incorporate the execution results into the reasoning chain.
Following~\citet{yao2022react}, involve multiple rounds of Thought-Action-Observation: 
\begin{equation}
\mathcal{H} = (s_0,a_0,o_0,s_1,\ldots,s_{t},a_{t},o_{t})
\end{equation}
where the agent articulates thoughts $s$, performs actions $a$ (Python), and receives observations $o$ as feedback from the interpreter. Action segments invoking python are closed by \actionl\ and \actionr\, more details are provided in Appendix~\ref{sec:appendix_steps_prompt}.

\item \textbf{Verdict.} In the final stage, the agent transitions from a local, step-wise analysis to a global, holistic assessment. It aggregates the evidence collected during the previous phase to issue a definitive judgment on the solution's correctness. The output is a strict binary classification (\raisebox{0.2ex}{\true} / \raisebox{0.2ex}{\false})  enclosed by \answerl\ and \answerr\, we adopt the logits of \true\ serving as the confidence signal for the verdict.
\end{itemize}

\paragraph{Backward Agent.}
The backward agent is designed to identify errors the forward agent may overlook, such as solutions that superficially logically sound but fail to satisfy problem constraints or omit necessary proof. It verifies the necessity of a solution by reasoning in reverse, from the final answer back to the problem statement.
Following the "Plan-Validate-Verdict" pipeline, the backward agent decomposes the solution into backward-checkable steps, systematically checks whether all required conditions are fulfilled.
We aggregate the results from forward and backward agents for a bidirectional and reliable assessment in Appendix~\ref{sec:appendix_aggregation}.

\begin{table*}[th]
\centering
\vspace{-2mm}
\resizebox{\textwidth}{!}{
\begin{tabular}{l|cccccccccccc}
\toprule[1.6pt]
    \textbf{Models} & \multicolumn{3}{c|}{\textbf{MATH500}} & \multicolumn{3}{c|}{\textbf{GSM8K}} & \multicolumn{3}{c|}{\textbf{Gaokao2023}} & \multicolumn{3}{c}{\textbf{AIME24}}  \\ \cmidrule{1-13}
    \textbf{BoN} & \textit{\textbf{@32}} & \textit{\textbf{@64}} & 
    \multicolumn{1}{c|}{\textit{\textbf{@128}}} & \textit{\textbf{@32}} 
    & \textit{\textbf{@64}} & \multicolumn{1}{c|}{\textit{\textbf{@128}}}  & \textit{\textbf{@32}}  & \textit{\textbf{@64}}  & \multicolumn{1}{c|}{\textit{\textbf{@128}}}  &  \textit{\textbf{@32}} & \textit{\textbf{@64}}  &  \textit{\textbf{@128}}   \\ 
    \midrule
        \multicolumn{13}{c}{\textit{\textbf{Text Reasoning LLM}}} \\
    \midrule
    Qwen2.5-7B-Instruct
& 53.0 & 50.4 & 51.0
& 85.5 & 85.7 & 83.9
& 40.8 & 42.3 & 41.3
& 30.0 & \underline{26.7} & \underline{33.3} \\

Llama3.1-8B-Instruct
& 46.7 & 48.0 & 45.2
& 81.5 & 80.1 & 81.7
& 34.5 & 35.6 & 35.6
& 30.0 & \textbf{30.0} & 30.0 \\

Qwen3-4B-Think
& \textbf{70.0} & \textbf{72.6} & \textbf{72.4}
& \textbf{92.3} & \textbf{91.2} & \textbf{92.2}
& \textbf{52.2} & \textbf{51.4} & \textbf{51.9}
& \textbf{40.0} & \textbf{30.0} & \textbf{36.7} \\

DS-Distill-14B
& \underline{66.2} & \underline{69.6} & \underline{69.4}
& \underline{90.1} & \underline{88.9} & \underline{89.8}
& \underline{48.3} & \underline{49.1} & \underline{48.6}
& \underline{33.3} & \underline{26.7} & \underline{36.7} \\

Mistral-Small-24B-Instruct
& 52.2 & 51.0 & 51.8
& 85.4 & 85.5 & 85.9
& 39.0 & 38.4 & 40.0
& 30.0 & \textbf{30.0} & \textbf{36.7} \\
    \specialrule{1pt}{0.4ex}{0.4ex}

    \multicolumn{13}{c}{\textit{\textbf{Outcome-level RM}}} \\
    \midrule

    GRM-Gemma-2B
& 45.6 & 48.8 & 46.6
& 77.9 & 75.5 & 73.4
& 33.5 & 31.9 & 33.2
& {33.3} & 26.7 & 30.0 \\

Skywork-V2-Llama-8B
& \underline{54.4} & \underline{55.2} & \underline{53.8}
& 87.5 & 87.3 & {87.6}
& \textbf{41.6} & 37.4 & {39.7}
& 30.0 & \underline{33.3} & \underline{36.7} \\

InternLM2-20B-RM
& {54.0} & {52.2} & {53.6}
& \underline{89.5} & \underline{89.6} & \underline{89.8}
& \textbf{41.6} & \textbf{42.9} & \underline{43.9}
& \underline{36.7} & \textbf{36.7} & {40.0} \\

INF-ORM-Llama3.1-70B
& \textbf{54.6} & \textbf{56.4} & \textbf{55.4}
& \textbf{91.2} & \textbf{90.8} & \textbf{91.5}
& \underline{40.8} & \underline{42.6} & \textbf{44.4}
& \textbf{40.0} & \textbf{36.7} & \textbf{40.0} \\

Starling-RM-34B
& 50.0 & 52.0 & 50.8
& {88.4} & {87.6} & 86.3
& 40.5 & {39.7} & 39.0
& 26.7 & 33.0 & \underline{36.7} \\

    \specialrule{1pt}{0.4ex}{0.4ex}

    \multicolumn{13}{c}{\textit{\textbf{Process-level RM}}} \\
    \midrule
    Qwen2.5-Math-PRM-7B
& \underline{65.2} & \textbf{69.4} & \textbf{70.2}
& \textbf{94.5} & \textbf{94.7} & \textbf{95.4}
& \textbf{50.9} & \textbf{51.4} & \textbf{54.3}
& \textbf{43.3} & \textbf{43.3} & \textbf{46.7} \\

Math-Shepherd-Mistral-7B-PRM
& \textbf{66.6} & \underline{66.6} & \underline{66.6}
& {90.0} & {89.2} & {89.6}
& {43.1} & {41.4} & {39.3}
& \textbf{30.0} & {36.7} & {30.0} \\

EurusPRM-Stage1
& 38.8 & 36.0 & 33.8
& 70.0 & 66.3 & 62.5
& 33.2 & 29.1 & 26.8
& 20.0 & 23.3 & 26.7 \\

EurusPRM-Stage2
& 36.0 & 31.8 & 31.0
& 66.0 & 62.1 & 57.5
& 29.1 & 26.5 & 25.7
& 16.7 & 20.0 & 23.3 \\

Skywork-PRM
& {61.6} & {66.2} & {65.6}
& \underline{92.5} & \underline{92.7} & \underline{93.4}
& \underline{46.0} & \underline{47.0} & \underline{46.8}
& {23.3} & \textbf{40.0} & \textbf{33.3} \\

    \specialrule{1pt}{0.4ex}{0.4ex} 
    \multicolumn{13}{c}{\textit{\textbf{Ours}}} \\
    \midrule
    Agentic-Verifier-Llama3-8B
    & 59.6 & 60.8 & 60.6
    & 90.8 & 91.2 & 90.7
    & 45.5 & 47.3 & 47.0
    & 30.0 & 33.3 & 40.0 \\
    
    Agentic-Verifier-Qwen3-4B
    & \textbf{73.8} & \textbf{76.2} & \textbf{79.0}
    & \textbf{93.0} & \textbf{92.6} & \textbf{93.3}
    & \textbf{54.5} & \textbf{55.1} & \textbf{57.4}
    & \textbf{46.7} & \textbf{50.0} & \textbf{53.3} \\

    \bottomrule[1.6pt]
\end{tabular}}
\caption{
\textbf{Performance (\%) of Best-of-N sampling on GSM8K, MATH500, Gaokao2023, and AIME24.} Within each category, the best results are in bold and the second-best results are underlined.
}
\label{Tab:Main_Exp_BoN}
\vspace{-5mm}
\end{table*}

\subsection{AgentV-RL}
The bidirectional design detailed above provides an effective multi-agent framework for identifying flaws in the candidate solution. 
Furthermore, to distill the capability of the multi-agent into a single LLM, a dedicated training recipe is necessary. 
In this section, we propose \textbf{AgentV-RL}, a scalable recipe comprising a synthetic data engine and a two-stage training schema.

\subsubsection{Synthetic Trajectories Sampling}
To bootstrap the verifier in performing tool-integrated verification, we design a synthetic data pipeline to obtain multi-turn high-quality trajectories. 
To achieve this, we curate the data by filtering high-quality public datasets, including Polaris~\citep{An:2025:polaris}, DeepScaleR-40K~\citep{Luo:2025:deepscaler}, and AReaL-boba-106k~\citep{Mei:2025:AReaL}.
For the diversity of synthetic data, we sample $k$ solutions for each question (setting $k=8$) and filter out questions(where $k$ solutions are all correct/incorrect). 
This yields an initial dataset containing solution candidates, $\mathcal{D}_{init} = \{x, y, l\}$, where $l \in \{\text{\raisebox{0.2ex}{\true}} / \text{\raisebox{0.2ex}{\false}}\}$ indicates the correctness of the solution $y$.

\par
Based on $\mathcal{D}_{init}$, we employ the LLM to automate the generation of tool-augmented verification trajectories. 
Specifically, LLM role plays either the forward or backward Agent, generating verification trajectory $\mathcal{H}$ that concludes with a final verdict $\tilde{l}$. 
Besides, we retain only those trajectories where the predicted verdict $\tilde{l}$ matches the ground truth $l$ for quality controlling. 
This results in the final dataset $\mathcal{D}_{\text{sft}}$, which comprises both positive and negative verification examples:
\begin{equation}
\mathcal{D}_{\text {sft }}=\left\{\left(x, y^{+}, \mathcal{H}^{+}\right)\right\} \cup\left\{\left(x, y^{-}, \mathcal{H}^{-}
\right)\right\},
\end{equation}
where the former term denotes the positive rollout $\mathcal{H}^{+}$ verified as \true, and the latter is negative rollout $\mathcal{H}^{-}$ validated as \false. 

\subsubsection{Rejection Fine-tuning}
With the synthetic dataset $\mathcal{D}_{\text{sft}}$, we conduct supervised fine-tuning (SFT) to endow the verifier with agentic proficiency. This stage is designed to align the model's behavior with the multi-turn, decision-making processes.

\par
The core objective is to encourage the verifier to reproduce stepwise reasoning and effective tool interaction. Formally, for each data point $(x, y, \mathcal{H})$, where $x$ is the problem, $y$ is the candidate solution, and the annotated verification trajectory $\mathcal{H}=(\tau_0,\tau_1,\ldots,\tau_n )$, where each $\tau_i \in \{s_i,a_i,o_i\}$, we optimize the following loss:
\begin{equation}
\mathcal{L} = - \mathbb{E}_{\tau \sim \mathcal{H}} \left[\sum_{i=1}^{|\mathcal{H}|} \mathbb{I}\left[\tau_i \neq o_i \right] \cdot \log \pi_\theta\left(\tau_i \mid \mathcal{H}_{<i}\right)\right]
\end{equation}

\subsubsection{Reinforcement Learning}
To further unlock the reasoning potential of the Agentic Verifier and incentivize autonomous exploration, we integrate Group Relative Policy Optimization (GRPO) within our recipe. Following the SFT phase, this stage aims to optimize the reasoning patterns to achieve multi-turn, long-horizon, and tool-integrated reasoning.
Specifically, we sample a group of candidate trajectories from verifier $\pi_{\psi}$ for each question-solution pair,
\begin{equation}
\left\{\mathcal{H}_i\right\}_{i=1}^G \sim \pi_{\psi}(\cdot \mid x, y, \mathbf{I}),
\end{equation}
where $\mathbf{I}$ represents the specific instruction prompt. In our experiments, we employ a mixed sampling strategy where $\pi_{\psi}$ acts as either forward agent or backward agent, allowing balanced optimization for dual-perspectives.

\section{Experiments}
\label{sec:exp}
\begin{table*}[t]
    \centering
    \vspace{-2mm}
    \resizebox{\textwidth}{!}{
    \begin{tabular}{l|cccccccccccc}
    \toprule
        \textbf{Verifier} & \multicolumn{3}{c}{\textbf{MATH500}} & \multicolumn{3}{c}{\textbf{GSM8K}} & \multicolumn{3}{c}{\textbf{Gaokao2023}} & \multicolumn{3}{c}{\textbf{AIME24}}  \\ \cmidrule{1-13}

        \textbf{Metrics} & \textit{\textbf{Acc}} & \textit{\textbf{$\Delta_\uparrow$}} & \textit{\textbf{$\Delta_\downarrow$}} & \textit{\textbf{Acc}} & \textit{\textbf{$\Delta_\uparrow$}} & \textit{\textbf{$\Delta_\downarrow$}}
        & \textit{\textbf{Acc}} & \textit{\textbf{$\Delta_\uparrow$}} & \textit{\textbf{$\Delta_\downarrow$}}
        & \textit{\textbf{Acc}} & \textit{\textbf{$\Delta_\uparrow$}} & \textit{\textbf{$\Delta_\downarrow$}}
        \\ 
        \midrule
            \multicolumn{13}{c}{\textit{\textbf{Turn 1}}} \\
        \midrule
        Qwen2.5-7B-Instruct
& 60.4 & 14.0 & 0.6
& 85.9 & 5.9 & 0.6
& 47.0 & 12.7 & 2.9
& 13.3 & 0.0 & 0.0 \\

Llama3.1-8B-Instruct
& 54.6 & 11.0 & 3.4
& 84.1 & 4.6 & 1.1
& 44.9 & 10.7 & 2.9
& 13.3 & 0.0 & 0.0 \\

Qwen3-4B-Think
& 80.0 & 33.8 & 0.8
& 90.9 & 11.1 & 0.9
& 64.9 & 31.2 & 3.4
& 16.7 & 3.3 & 0.0 \\

DS-Distill-14B
& \underline{83.0} & 37.4 & 1.4
& \underline{92.0} & 12.1 & 0.8
& \underline{67.0} & 32.5 & 2.6
& \underline{23.3} & 10.0 & 0.0 \\

Mistral-Small-24B-Instruct
& 61.4 & 15.6 & 1.2
& 88.4 & 8.5 & 0.8
& 49.1 & 15.6 & 3.6
& 13.3 & 0.0 & 0.0 \\

Agentic-Verifier-Qwen3-4B
& \textbf{84.2} & 41.6 & 0.6
& \textbf{94.6} & 14.5 & 0.5
& \textbf{75.6} & 40.3 & 1.8
& \textbf{40.0} & 26.7 & 3.3 \\
        \midrule
            \multicolumn{13}{c}{\textit{\textbf{Turn 2}}} \\
        \midrule
       Qwen2.5-7B-Instruct
& 64.8 & 6.2 & 1.8
& 87.4 & 3.3 & 1.9
& 49.9 & 5.2 & 2.3
& 6.7 & 0.0 & 6.7 \\

Llama3.1-8B-Instruct
& 58.4 & 6.0 & 2.2
& 82.8 & 3.3 & 4.6
& 48.1 & 5.2 & 2.1
& 10.0 & 0.0 & 3.3 \\

Qwen3-4B-Think
& 84.6 & 6.0 & 1.4
& 92.2 & 2.4 & 1.1
& 69.1 & 5.5 & 1.3
& 23.3 & 6.7 & 3.3 \\

DS-Distill-14B
& \underline{87.4} & 5.8 & 1.4
& \underline{92.7} & 2.4 & 1.7
& \underline{70.1} & 4.7 & 1.6
& \underline{26.7} & 3.3 & 0.0 \\

Mistral-Small-24B-Instruct
& 66.4 & 5.8 & 0.8
& 90.0 & 2.7 & 1.1
& 52.7 & 4.9 & 1.3
& 13.3 & 0.0 & 0.0 \\

Agentic-Verifier-Qwen3-4B
& \textbf{89.2} & 2.4 & 1.2
& \textbf{94.1} & 0.4 & 0.9
& \textbf{76.6} & 3.4 & 2.3
& \textbf{33.3} & 0.0 & 6.7 \\
        \midrule
            \multicolumn{13}{c}{\textit{\textbf{Turn 3}}} \\
        \midrule
        Qwen2.5-7B-Instruct
& 66.0 & 4.2 & 3.0
& 87.5 & 2.96 & 2.8
& 52.2 & 3.64 & 1.3
& 10.0 & 6.7 & 3.3 \\

Llama3.1-8B-Instruct
& 58.8 & 4.0 & 3.6
& 82.8 & 3.3 & 3.3
& 47.0 & 1.3 & 2.3
& 6.0 & 0.0 & 3.00 \\

Qwen3-4B-Think
& 84.8 & 2.4 & 2.2
& 92.0 & 1.2 & 1.4
& 69.6 & 2.3 & 1.6
& 26.0 & 6.7 & 3.0 \\

DS-Distill-14B
& \underline{85.8} & 2.2 & 3.8
& \underline{93.4} & 1.8 & 1.1
& \underline{73.0} & 4.4 & 1.6
& \underline{33.0} & 6.7 & 0.0 \\

Mistral-Small-24B-Instruct
& 67.8 & 4.4 & 3.0
& \cellcolor{lightgreen1}{90.6} & 2.65 & 2.05
& 52.2 & 2.08 & 2.60
& 13.0 & 0.00 & 0.00 \\

Agentic-Verifier-Qwen3-4B
& \textbf{89.8} & 1.8 & 1.2
& \textbf{94.1} & 1.0 & 1.0
& \textbf{76.4} & 2.6 & 2.9
& \textbf{33.0} & 3.3 & 3.3 \\
        \bottomrule[1.6pt]
    \end{tabular}
    }
\caption{\textbf{Performance comparison against different verifiers for iterative refinement.}
\textit{Acc} denotes the average accuracy at each refinement iteration, together with $\Delta_\uparrow$ and $\Delta_\downarrow$, which measure the rates of correction and degradation relative to the previous iteration.
The best result is in \textbf{bold}, and the second-best is \underline{underlined}.
}
\label{Tab:Main_Exp_Refine}
\vspace{-2mm}
\end{table*}

\paragraph{Verifiable Reward.}
We apply a binary outcome-level reward by comparing the consistency between the predicted verdict $\tilde{l}$ and ground truth $l$,
\begin{equation}
r(\mathcal{H})= \begin{cases}1, & \text { if  } \tilde{l}=  l 
 \\ -1, & \text { otherwise } \end{cases}
\end{equation}
Following~\citet{Yu:2025:DAPO}, we dynamically filter those zero reward-variance groups of rollouts, which is too easy or hard for policy model (predominantly all +1 or all -1). Then derive the normalized advantages $\hat{A}_{i, t}$ within each group and update the verifier with the objective:
\begin{align}
& \mathcal{J}_{\mathrm{GRPO}}(\psi)=\mathbb{E}_{(x,y ) \sim \mathcal{D},\left\{\mathcal{H}_i\right\}_{i=1}^G \sim \pi_{\psi_{\text {old }}}} 
\frac{1}{G}  \sum_{i=1}^G \frac{1}{\left|\mathcal{H}_i \right|} \nonumber  \\
& \sum_{t=1}^{\left| \mathcal{H}_i \right|}  
\min( r_{i, t}(\psi) \hat{A}_{i, t}, \operatorname{clip} (r_{i, t}(\psi), 1-\epsilon_{low}, 1+  \nonumber  \\
& \epsilon_{high}) \hat{A}_{i, t} ) - \beta D_{\mathrm{KL}}\left(\pi_\psi \| \pi_{\text {ref }}\right) 
\end{align}
where $r_{i, t}(\psi)=\frac{\pi_\psi\left(\mathcal{H}_{i} \mid x, y, \mathcal{H}_{i,<t} \right)}{\pi_{\psi_{\text {old }}}\left(\mathcal{H}_{i} \mid x, y, \mathcal{H}_{i,<t}\right)}$ is the importance sampling ratio.
To mitigate the risk of memorizing environment observations, we exclude the compiler execution from the loss calculation. 

\subsection{Setup}
\paragraph{Task.}
We evaluate the proposed method from two popular TTS paradigms: parallel scaling and sequential scaling. The former focuses on BoN, while the latter is verifier-revision iteratively. 
We conduct experiments on widely used reasoning benchmarks: GSM8K~\citep{Cobbe:2023:GSM8k}, AIME24~\citep{AIME:2024:AIME}, MATH500~\citep{Lightman:2024:MATH500} and Gaokao2023~\citep{Liao:2024:Gaokao2023}.

\paragraph{Implementation Details.}
We utilize Qwen3-4B~\citep{Yang:2025:Qwen3} as our base model. In the SFT phase, the verifier is trained on 15K synthesized samples constructed in Section~\ref{sec:method}. The model is subsequently further optimized using GRPO on an additional 50K samples. We provide implementation illustrations in Appendix~\ref{sec:appendix_impl_details}.

\paragraph{Evaluation Setting.}
For parallel scaling, we report the accuracy under different sampling rollouts $N$, where the verifier scores $N$ candidate solutions and selects the highest-scoring one as the final output. For sequential scaling, we report the average accuracy at each refinement iteration, together with $\Delta_\uparrow$ and $\Delta_\downarrow$, which measure the rates of correction and degradation relative to the previous iteration. To ensure strict comparability, all verifier variants are evaluated on the same fixed candidate pool and the same initial solution for each problem instance. Additional evaluation details and baseline descriptions are provided in the  Appendix~\ref{sec:appendix_eval_protocol} and \ref{sec:appendix_baselines}.

\subsection{Main Result}
\captionsetup{skip=4pt}
\begin{figure*}[t]
\centering
\begin{subfigure}[t]{0.49\textwidth}
\centering
\includegraphics[width=\linewidth]{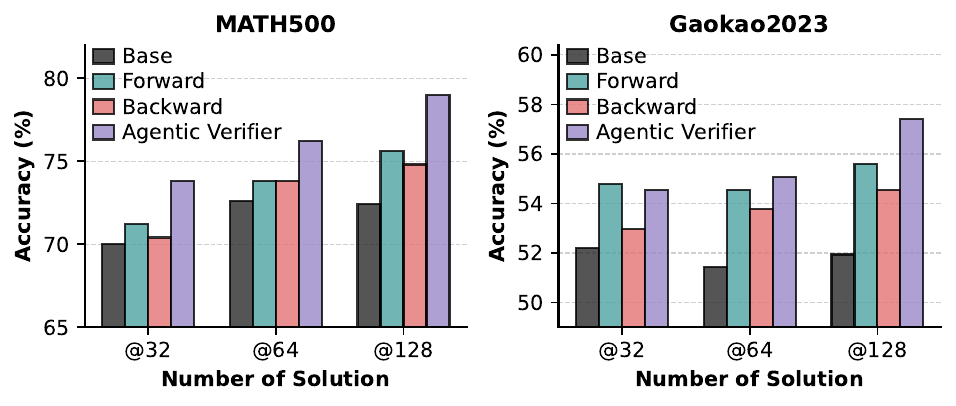}
\label{fig:bon}
\end{subfigure}
\hfill
\begin{subfigure}[t]{0.49\textwidth}
\centering
\includegraphics[width=\linewidth]{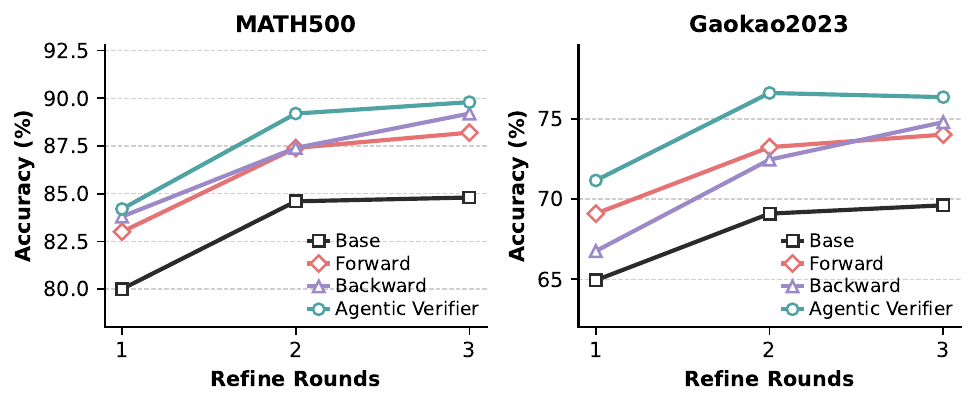}
\label{fig:refine}
\end{subfigure}
\vspace{-10pt}
\caption{
\textbf{Ablation study of different variants on Best-of-N sampling (left) and verifier revision (right).}
Forward-only and backward-only variants are both competitive, while the full design performs best.
}
\vspace{-5mm}
\label{fig:ablation_bon_refine_component}
\end{figure*}

\paragraph{Agentic Verifier improves Best-of-N sampling performance.}
Table~\ref{Tab:Main_Exp_BoN} reports BoN performance across state-of-the-art competitors. 
We make several key observations:
\textit{(1)}~Agentic Verifier delivers consistently strong results across all benchmarks, establishing new state-of-the-art performance. Notably, Agentic-Verifier-Qwen3-4B achieves the highest accuracy on MATH500 (up to 79.0\%), surpassing the previous best outcome-level RM, Skywork-V2-Llama-8B, by a substantial margin of 25.2 percentage points. Similar trends are observed for GSM8K, Gaokao2023, and AIME24, with our model maintaining significant leads.
\textit{(2)}~Importantly, as the number of BoN samples $N$ increases (from 32 to 128), Agentic Verifier’s performance continues to improve steadily, particularly on challenging benchmarks like AIME24. On AIME24, the 4B variant reaches 53.3\% accuracy at $N=128$, while all baselines—including larger models—remain significantly lower. 
\textit{(3)}~Our variants deliver a substantial improvement compared to the base model, e.g. Qwen3-4B and Llama-3.1-8B.
The substantial gains stem from the bidirectional verification and seamless tool integration in the proposed framework, enabling it to scrutinize reasoning chains and catch subtle errors—especially for challenge benchmarks like AIME24.

\paragraph{Agentic Verifier provides effective guidance for iterative refinement.}
To assess the helpfulness of the verifiers' feedback, we conduct a thorough analysis of critique-revision strategies with different verifiers. 
Based on the results in Table~\ref{Tab:Main_Exp_Refine}, we can conclude:
\textit{(1)}~In the 1-st turn, the actor achieves 94.6\% accuracy on GSM8K and 84.2\% on MATH500 with the critique from Agentic-Verifier-Qwen3-4B.
As indicated by the up to 41.6\%  $\Delta_\uparrow$ (percentage of wrong solutions corrected), while maintaining a minimal rate 0.6\% of incorrectly revised solutions ($\Delta_\downarrow$).
Agentic Verifier delivers high-quality critiques that assist the actor in correcting erroneous solutions. 
\textit{(2)}~Compared to other verifiers, Agentic Verifier achieves faster convergence: significant performance boosts are observed in the first one or two refinement rounds. 
The performance remains stable or only slightly declines in later iterations, avoiding the degradation encountered like DS-Distill-14B. This robustness can be attributed to its agentic multi-turn reasoning and tool-augmented verification mechanisms, which allow it to maintain rigorous oversight throughout the refinement process.

\subsection{Analysis}
\subsubsection{Ablation Study}
\paragraph{Effectiveness of Bidirectional Components.}
Recall that Agentic Verifier comprises two specialized agents: a forward agent that traces the logical flow from premises to conclusion within the candidate solution, and a backward agent that verifies necessity by reasoning from the final conclusion back to the original query.
To assess the individual contributions of each agent, we conduct an ablation study comparing two variants on both BoN sampling and verifier-revision, as presented in Fig.~\ref{fig:ablation_bon_refine_component}: (a) forward agent only, and (b) backward agent only. Both variants achieve competitive results; however, Agentic Verifier, which integrates both agents, achieves the state-of-the-art performance. This demonstrates that our bidirectional design objectives are synergistic, leading to improved generalization and reliable verification.

\paragraph{Effectiveness of Tool Usage.}
We observe that even the variant without tool already significantly outperforms the base model, indicating that the Agentic Verifier itself contributes substantially to performance gains, beyond what is achieved by tool alone. When the tool is incorporated, performance improves further. Notably, tool usage remains moderate and stable in practice. For a detailed analysis, including ablations and execution statistics, please refer to Appendix~\ref{sec:appendix_tool}.

\begin{figure}[t]
\centering
\includegraphics[width=\linewidth]{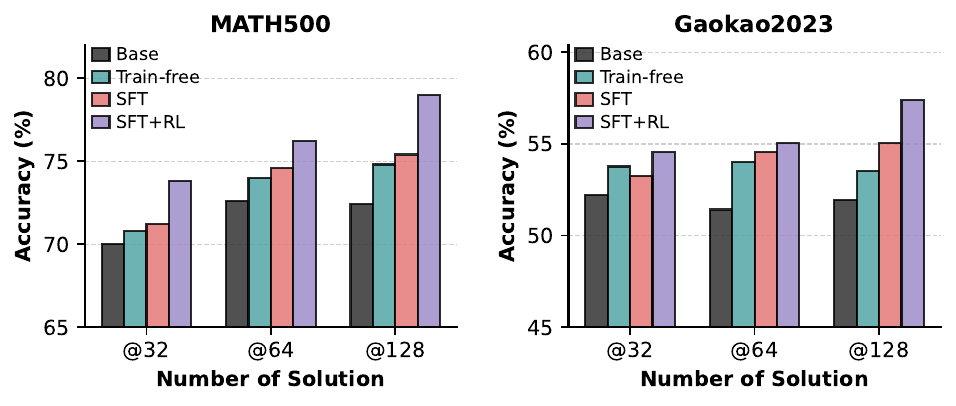}
\caption{
\textbf{Controllable study of different training design choices on BoN performance.}
The SFT+RL design achieves the strongest results.
}
\vspace{-5mm}
\label{fig:bon_training}
\end{figure}

\paragraph{Training Recipe.}
We explore the key ingredients in our training recipe.
Through a series of controllable studies, we quantitatively evaluate the effect of our main design choices by comparing the following configurations: Train-free, SFT, and SFT+RL. 
Figure~\ref{fig:bon_training} presents the results of our experiments and more details about settings can be found in Appendix~\ref{sec:appendix_impl_details}.
We observe that the Train-free variant already achieves relatively competitive results, outperforming the base model by up to 2.6  points on Gaokao2023. 
Furthermore, the SFT variant demonstrates additional performance gains, confirming the effectiveness of our synthetic data engine. 
Finally, the introduction of RL further unlocks the model’s reasoning potential by enabling direct interaction with the environment and optimizing behavioral patterns.

\subsubsection{Scaling Effect}
\begin{figure}[t]
\centering
\includegraphics[width=\linewidth]{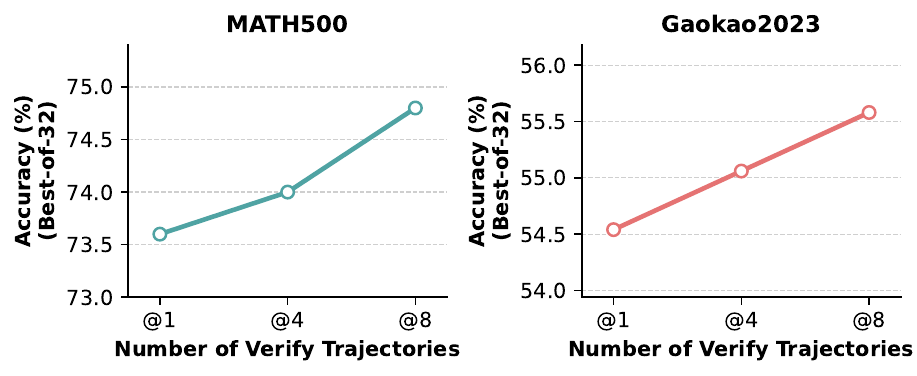}
\caption{
\textbf{Scaling inference-time compute for verification.}
Sampling multiple verification trajectories improves BoN accuracy.
}
\label{fig:inference_time_scaling}
\vspace{-1mm}
\end{figure}

\paragraph{Scaling Inference-time Compute.}
Scaling inference-time compute within Agentic Verifier can be achieved by sampling multiple verification trajectories and aggregating their scores, as defined in Equation~\ref{eq:boN_TTS}.
Figure~\ref{fig:inference_time_scaling} demonstrates that BoN performance scales positively with the allocation of additional inference-time compute.

\paragraph{Scaling Model Size.}
We also examine the effect of scaling model size under train-free setting. As shown in Table~\ref{tab:bon_model_scale}, larger models consistently achieve higher accuracy across benchmarks. Scaling from 0.6B to 1.7B parameters yields an improvement of 5.2 points on Gaokao2023, with further gains observed at Qwen3-4B.
These findings demonstrate that scaling up model capacity continually benefits from our agentic framework. 

\begin{table}[t]
\centering
\small
\setlength{\tabcolsep}{6pt}
\begin{tabular}{l|ccc|ccc}
\toprule
 & \multicolumn{3}{c|}{\textbf{MATH500}} & \multicolumn{3}{c}{\textbf{Gaokao2023}} \\
\textbf{Model} & @32 & @64 & @128 & @32 & @64 & @128 \\
\midrule
0.6B & 59.8 & 63.2 & 62.2 & 43.9 & 44.9 & 43.1 \\
1.7B & 64.8 & 67.4 & 68.0 & 46.5 & 49.4 & 48.3 \\
4B   & \textbf{73.8} & \textbf{76.2} & \textbf{79.0} 
     & \textbf{54.5} & \textbf{55.1} & \textbf{57.4} \\
\bottomrule
\end{tabular}
\caption{
\textbf{Scaling model size of Agentic Verifier with the Qwen3 series.}
Larger models consistently benefit from the proposed agentic verification framework.}
\vspace{-1mm}
\label{tab:bon_model_scale}
\end{table}

\subsubsection{Generalization Performance}
We further evaluate Agentic Verifier on LiveCodeBench~\cite{jain2024livecodebench} and HotpotQA~\cite{yang2018hotpotqa} to test its versatility on competitive code and multihop QA. 
Agentic-Verifier-Qwen3-4B achieves superior performance on both benchmarks (Table~\ref{tab:generalization}), confirming that our bidirectional, tool-augmented paradigm generalizes to broader reasoning domains. See Appendix~\ref{sec:beyond_math} for experimental details.

\begin{table}[t]
\centering
\small
\setlength{\tabcolsep}{10pt}
\begin{tabular}{lcc}
\toprule
Model & LCB & HotpotQA \\
\midrule
Qwen3-4B & 57.14 & 40.00 \\
Qwen2.5-7B-Instruct & 58.29 & 40.67 \\
DS-Distill-14B & 64.57 & 52.05 \\
Mistral-Small-24B-Instruct & 50.30 & 63.00 \\
Agentic-Verifier-Qwen3-4B & \textbf{70.86} & \textbf{66.00} \\
\bottomrule
\end{tabular}
\caption{\textbf{Generalization Performance} on LiveCodeBench and HotpotQA.}
\label{tab:generalization}
\end{table}

\subsubsection{Latency Analysis}
\paragraph{Computational Overhead.} As detailed in Table~\ref{tab:latency}, Agentic Verifier achieves the highest accuracy at the expense of increased tokens and latency. This overhead stems from its multi-turn deliberative process, yet the Forward and Backward variants provide scalable trade-offs for resource-constrained settings. Benchmark results are averaged over trajectories using batch size of 128 on A100 GPU.

\begin{table}[t]
\centering
\small
\setlength{\tabcolsep}{3.8pt}
\begin{tabular}{lcccc}
\toprule
Model & Tokens & Rounds & Tool Calls & Time (s) \\
\midrule
Base & 2560 & 1.0 & 0.0 & 119.0 \\
Forward & 4114 & 5.7 & 0.9 & 159.1 \\
Backward & 4235 & 5.6 & 0.7 & 164.3 \\
Agentic-Verifier & 8349 & 11.3 & 1.6 & 323.4 \\
\bottomrule
\end{tabular}
\caption{
\textbf{Computational cost and latency analysis.}
All metrics are averaged over verification trajectories, and latency is measured on a single A100 GPU using vLLM with batch size 128.
}
\label{tab:latency}
\vspace{-2mm}
\end{table}

\section{Conclusion}
This paper introduces Agentic Verifier, an agentic reward modeling paradigm that leverages multi-turn, tool-augmented verification. 
By orchestrating forward and backward agents, our approach enables comprehensive and interpretable solution checking.
Extensive experiments show that Agentic Verifier achieves substantial improvements in both parallel and sequential TTS, outperforming SOTA ORM and PRM. These results highlight Agentic Verifier as a promising direction for advancing agentic reward modeling.
We hope our work provides meaningful insights for the development of reward models and inspires future research on autonomous verification and reasoning.

\section{Limitations}
Though Agentic Verifier displays state-of-the-art effectiveness, several limitations should be noted. First, the reliance on synthetic, tool-augmented data may not fully represent the variety of real-world reasoning problems, which impede generalization. Second, the multi-turn process increases computational cost, presenting challenges for real-time or resource-limited deployment. Finally, the framework’s performance is tied to the coverage and reliability of external tools, which can be a bottleneck for certain tasks. 
Future work will focus on optimizing efficiency, broadening tool integration, and improving adaptability to a wider range of tasks.

\section{Ethics Statement}
This paper introduces an agentic reward modeling paradigm designed to enhance the reliability of automatic solution verification. 
The methods and findings presented are intended solely for research purposes; Our study does not pertain to malicious applications, unintended uses, or issues related to fairness, privacy, security, crowdsourcing, or human subject research.

\section*{Acknowledgments}
The authors wish to thank the anonymous reviewers for their helpful comments. This work was partially funded by National Key R\&D Program of China No.2025ZD0215702, National Natural Science Foundation of China (No.~162576106, 62476061, 62376061).

\bibliography{anthology}

\appendix

\section{Use Of AI Assistants}
LLMs are utilized to polish the article to improve the reading experience. 

\section{Details of Evaluation}
\label{sec:appendix_prompts}

\subsection{Text Reasoning Prompt}
\label{sec:appendix_baseline_prompts}

For reference, we include two commonly used prompting strategies as baselines. The Best-of-N verifier evaluates each candidate solution
independently in a single turn and outputs a boolean correctness judgment. The Refine-style verifier re-evaluates an initial judgment by providing
feedback conditioned on the original problem and solution.

\subsubsection{Best-of-N}
\label{sec:appendix_bon_prompt}

\vspace{2pt}
\noindent\textbf{System}
\begin{promptbox}
You are a teacher. Your task is to grade the solution, verifying correctness. 
Use Expected Answer to find any erroneous step in the Solution.

## GOAL

Given a problem and a solution, you must:

1) write a concise verification text that inspects the given solution step-by-step (do not re-solve unless a single local derivation is trivially needed);

2) decide whether the solution correctly solves the problem;

3) output a boolean verdict.

You must prioritize checking the original solution via paper checks: legality of algebraic steps, substitution mentally for proposed roots, domain and edge cases, theorem prerequisites, and consistency of the final statement with intermediate steps. Do NOT produce a fresh full solution when the provided reasoning is wrong or incomplete.

## ALLOWED TAGS

* <rubric>...</rubric>  - required at the start; list 2 - 4 decisive axes you'll check.
* <think>...</think>    - private reasoning exactly once; each update must add evidence or tighten the verdict (state micro-goal, the two most decisive axes, a compact known/unknown ledger, then choose the smallest next step: conclude or one precise mental check).
* <verify>...</verify>  - public verification text, focused on checking the given reasoning.
* <answer>...</answer>  - final boolean verdict (true/false) exactly once.

\end{promptbox}

\vspace{2pt}
\noindent\textbf{User}
\begin{promptbox}
### Problem ###

{problem}

### Solution ###

{solution}

### Verification Task Reminder ###

- Start with <rubric>...</rubric>.
- Include exactly <think>...</think>.
- Then verify the solution step by step and output the final verdict:<verify>...</verify><answer>true|false</answer>
\end{promptbox}

\subsubsection{Refine}
\label{sec:appendix_refine_prompt}

\vspace{2pt}
\noindent\textbf{System}
\begin{promptbox}
You are a Verifier agent responsible for performing a verification of a math problem's solution.Your mission is to determine whether the given solution is correct.
\end{promptbox}

\vspace{2pt}
\noindent\textbf{User}
\begin{promptbox}
Read the following carefully and think critically:

### Problem ###

{problem}

### Solution ###

{solution}

Your are required to verify the solution carefully and show your reasoning process.
At the end, provide your final judgment in a single line using:

<answer>true</answer>  - if the reasoning confirms that the conclusion logically holds.

<answer>false</answer> - if there exists any missing justification, invalid reversal, or unsupported dependency.
\end{promptbox}

\subsection{Agentic Verifier Prompt}
\label{sec:appendix_ours_prompts}

For reference, we report the prompt templates used in our proposed
multi-stage verification framework.
The framework decomposes verification into three stages, namely
Plan, Validation, and Verdict.
We consider two planning variants: a forward verifier, which plans
verification steps by reasoning from the original problem, and a
backward verifier, which plans by reasoning backward from the final
conclusion.
Both variants share the same step-wise Validation and Verdict prompts.
Intermediate user prompts that advance the verifier between stages are
automatically inserted by the system during execution.

\subsubsection{System Prompt}

\textbf{Forward}
\begin{promptbox}
You are a Verifier agent responsible for performing a multi-turn verification of a math problem's solution.

Your mission is to determine whether the given solution is correct.

You must complete the task strictly in three distinct stages, executed sequentially and independently.

## Stage A: Task Analysis \& Extraction

Analyze the original question and its provided solution. Decompose the verification into smaller, checkable steps.

## Stage B: Solution Analysis \& Judgment

Perform the actual verification one step at a time, following the plan you created in Stage A.

## Stage C: Final Review \& Verdict

After all Stage B steps are complete, review the entire reasoning process and produce a final boolean verdict.

In every turn, you must include your reasoning process inside a <think>...</think> block before presenting your results.
\end{promptbox}
\noindent\textbf{Backward}
\begin{promptbox}
You are a Verifier agent responsible for performing a multi-turn verification of a math problem's solution.
Your mission is to determine whether the given solution is correct.

Unlike a forward, step-by-step verifier, your reasoning proceeds in reverse: 
you begin from the final result or conclusion presented in the solution 
and trace backward to determine whether it can be rigorously justified by the given question and established facts.

You must complete the task strictly in three distinct stages, executed sequentially and independently.

## Stage A: Task Analysis \& Extraction

Analyze the original question and its provided solution. Decompose the verification into smaller, checkable backward steps.

## Stage B: Solution Analysis \& Judgment

Perform the actual verification one step at a time, following the plan you created in Stage A.

## Stage C: Final Review \& Verdict

After all Stage B steps are complete, review the entire reasoning process and produce a final boolean verdict.

In every turn, you must include your reasoning process inside a <think>...</think> block before presenting your results.
\end{promptbox}

\subsubsection{Stage A: Planning Prompts}
\label{sec:appendix_plan_forward}

\textbf{Forward}
\begin{promptbox}
You are now entering **Stage A: Task Analysis \& Extraction**.

### Problem ###

{problem}

### Solution ###

{solution}

In this stage, you are limited to analysis and planning only.
You must NOT verify, compute, or decide correctness during Stage A.

Your objectives:

* Break down the original question into its essential components.

* Decompose the provided solution into its main steps.

* Design a sequence of verification steps for Stage B.

  - Each verification step must describe *what* to check later, not perform the check itself.

  - Focus on logical reasoning, consistency, assumptions, and calculations that need to be validated in Stage B.

After you list all the planned verification steps, stop your generation immediately to indicate that Stage A is complete.
\end{promptbox}

\noindent\textbf{Backward}
\begin{promptbox}
You are now entering **Stage A: Backward Task Analysis \& Extraction**.

### Problem ###

{problem}

### Solution ###

{solution}

In this stage, you are limited to analysis and planning only.

You must NOT perform explicit verification, detailed calculations, or decide correctness during Stage A.

Your objectives:

* Identify the final conclusion or result asserted by the solution.

* Work backward from this conclusion to determine which intermediate statements or transformations must hold.

* Break down the original question into its essential components that support these backward dependencies.

* Decompose the provided solution into its main steps in a way that makes backward checking possible.

* Design a sequence of verification steps for Stage B.

  - Each verification step must describe *what* backward relationship needs to be checked later, not perform the check itself.
  
  - Focus on logical implications, hidden assumptions, reversibility of operations, and calculations that need to be validated in Stage B.

After you list all the planned verification steps, stop your generation immediately to indicate that Stage A is complete.
\end{promptbox}

\subsubsection{Stage B: Validation Prompt}
\label{sec:appendix_steps_prompt}

\begin{promptbox}
You are now entering **Stage B: Solution Analysis \& Judgment**.

In this stage, you will conduct a multi-turn verification of the steps designed in Stage A.
Each verification step may require one or more turns to complete.

**Turn Logic:**

- A single verification step may take multiple turns.

- Each turn must end with one of the following:

  * <python>...</python> - when you need to perform a computation before continuing the same step in the next turn.
  
  * <step/> - when the current step has been fully verified and you are ready to move to the next one.
  
  * <end\_of\_analysis/> - when all steps are verified or a clear error has been found.

**Instructions for each turn:**

1. Restate the step you are verifying and reason carefully about it.

2. Check whether this step is logically consistent with all previously verified steps.

3. If a calculation or detailed check is needed, output a <python>...</python> block after reasoning and stop your response.  
   The system will execute it and return the results for you to continue in the next turn.

**Rules for <python> blocks:**

- Use only necessary imports and print() statements for outputs.

- Do not use OS commands, file I/O, input(), or networking.

- A <python> block always marks the end of the current turn.

- Python tool can only be invoke at most three times across stage B.

\end{promptbox}

\subsubsection{Stage C: Verdict Prompt}
\label{sec:appendix_review_prompt}

\begin{promptbox}
Now you are required to conduct stage C.

Given all prior analyses, provide your final review and boolean verdict.

Requirements:
- Review all previous verification steps and summarize why each step was correct or incorrect in <review>...</review>.

- If all previous steps were confirmed to be correct, output <answer>true</answer>.

- If any step contained errors, or if you identify new inconsistencies at this stage, output <answer>false</answer>.
\end{promptbox}

\section{Experimental Details}

\subsection{Evaluation Protocol Details}
\label{sec:appendix_eval_protocol}

To ensure strict comparability across settings, all candidate solutions in both BoN and sequential refinement are generated by the same fixed actor, Qwen2.5-7B-Instruct, using identical sampling hyperparameters (temperature $=1.0$, top-$k=50$, and maximum completion length $=4096$). For each problem, we sample 128 rollouts once and reuse this shared candidate pool across all verifier variants. In BoN, every verifier evaluates exactly the same 128 rollouts. In sequential refinement, the initial solution is sampled from this same pool and kept fixed across all refinement variants.

\subsection{Baselines.}
\label{sec:appendix_baselines}

We compare against three families of baselines that are commonly used in
solution verification and preference modeling: (i) \emph{text-reasoning} judges
that directly generate a judgment from raw text, (ii) \emph{outcome reward models}
(ORMs) that score entire solutions, and (iii) \emph{process reward models}
(PRMs) that provide step-level supervision.

\paragraph{Text-Reasoning LLMs.}
These baselines use an instruction-tuned model as a text-only judge to produce
a free-form verification and/or a final correctness decision without any
explicit training.
We instantiate this family with the following models: Qwen2.5-7B-Instruct~\cite{Qwen2.5_Models}, Llama-3.1-8B-Instruct~\cite{Llama3_Models}, Qwen3-4B~\cite{Qwen3_Models}, DeepSeek-R1-Distill-Qwen-14B~\cite{Deepseek3_Models} and Mistral-Small-24B-Instruct~\cite{mistral-small-3-24b}.

\paragraph{Outcome Reward Models (ORM).}
ORMs assign a scalar score to a complete solution (or a response) to reflect
its overall quality/correctness under preference supervision.
We evaluate the following baselines: GRM-Gemma-2B~\cite{Gemma_2b_RM}, Skywork-V2-Llama-8B~\cite{Skywork_ORM_V2}, InternLM2-20B-RM~\cite{InternLM2}, INF-ORM-Llama3.1-70B~\cite{INF_ORM_Llama3.1_70B} and Starling-RM-34B~\cite{Starling-RM}.

\paragraph{Process Reward Models (PRM).}
PRMs provide step-level feedback along a reasoning trajectory, enabling process supervision beyond outcome-only scoring. We include state-of-the-art PRMs: Qwen2.5-Math-PRM-7B~\cite{Qwen_Math_PRM}, Math-Shepherd-Mistral-7B-PRM~\cite{Math-Shepherd-PRM}, EurusPRM~\cite{EurusPRM} and Skywork-PRM~\cite{Skywork_PRM}.

\subsection{Aggregation for Forward \& Backward Agents}
\label{sec:appendix_aggregation}
This section clarifies how verification results of forward and backward agents are aggregated in different experimental settings.

\paragraph{BoN.}
We average the confidence score from the forward and backward traces to obtain the aggregated score.
Candidate solutions are then ranked according to the aggregated score, and the highest-scoring solution is selected.
This score-based aggregation is sufficient for selection and introduces no additional modeling complexity.

\paragraph{Verifier Revision.}
In the multi-round verifier-revision, the forward and backward verification traces each produce a correctness judgment together with natural-language feedback.
We adopt a conservative aggregation rule: a solution is considered correct only if both verification traces judge it as correct.
If either verification trace identifies an error, the solution is regarded as incorrect and refinement is triggered. In this case, we invoke the verifier  once more to generate targeted modification suggestions.
The resulting feedback is then used as the supervisory signal for actor LLM.

\subsection{Experiment Setup}
\label{sec:appendix_impl_details}

This section summarizes the experimental training setup used in our study, including data construction, supervised fine-tuning, and reinforcement learning.

\paragraph{Data Construction.}
We construct our training data through a two-stage synthesis and filtering pipeline.
Qwen2.5-7B-Instruct is first used as a generator to sample question--answer pairs from diverse mathematical benchmarks.
We then employ Qwen3-4B as a vanilla verifier and perform a Best-of-8 verification for each sample, using the number of correct candidates as a coarse difficulty signal.
Based on this signal, the data are grouped by difficulty and used to support subsequent SFT and RL training.

\paragraph{Supervised Fine-Tuning.}
All models are fine-tuned using a learning rate of $5\times10^{-6}$ with a batch
size of 128 for a single epoch.
The maximum sequence length is set to 21{,}600 tokens.
During training, user inputs and tool outputs are masked and excluded from loss
computation.
SFT is conducted on 16 NVIDIA A100 GPUs (80GB) with bfloat16 precision enabled.

\paragraph{Reinforcement Learning.}
For reinforcement learning, we adopt Group Relative Policy Optimization (GRPO).
The actor model is trained with a learning rate of $1\times10^{-6}$, a per-device
batch size of 1, and gradient accumulation over 64 steps.
For each query, 8 candidate responses are sampled.
Each rollout uses a maximum completion length of 4{,}096 tokens per turn, with a
sampling temperature of 1.0, and the total rollout length is capped at 21{,}600 tokens.
The same masking strategy as in SFT is applied, excluding user inputs and tool
outputs from loss computation.
The number of tool invocations is limited to at most three per rollout.
Reinforcement learning is conducted on 32 NVIDIA A100 GPUs (80GB) with bfloat16
precision enabled.

\paragraph{Scaling Inference-time Compute for Verifier.}
For the same question $x$ and candidate answer $y$, the verifier samples $k$ verification
trajectories and aggregating their scores, as
\begin{align}
l_{\text{Agg@K}} &=\frac{1}{k}\sum_{i=1}^{k}\pi_{\psi}(\text{\true} \mid x, y, f_{i},\mathbf{I}) \nonumber   \\
f_{i}\sim &\pi_{\psi}( \cdot | x,y, \mathbf{I}) 
\label{eq:boN_TTS}
\end{align}

\subsection{Additional Ablation on Tool Usage}
\label{sec:appendix_tool}

\begin{table}[t]
\centering
\small
\begin{tabular}{lccc}
\toprule
Model & @32 & @64 & @128 \\
\midrule
Base & 70.60 & 72.60 & 72.40 \\
Agentic Verifier w.o. Tool & 73.20 & 76.00 & 77.40 \\
Agentic Verifier w. Tool & \textbf{73.80} & \textbf{76.20} & \textbf{79.00} \\
\bottomrule
\end{tabular}
\caption{\textbf{Tool grounding on Math500.} BoN accuracy under different sampling budgets.}
\label{tab:tool_ablation_math}
\end{table}

\begin{table}[t]
\centering
\small
\begin{tabular}{lccc}
\toprule
Model & @32 & @64 & @128 \\
\midrule
Base & 52.20 & 51.42 & 51.94 \\
Agentic Verifier w.o. Tool & 51.95 & 54.52 & 55.06 \\
Agentic Verifier w. Tool & \textbf{54.54} & \textbf{55.06} & \textbf{57.40} \\
\bottomrule
\end{tabular}
\caption{\textbf{Tool grounding on Gaokao2023.} BoN accuracy under different sampling budgets.}
\label{tab:tool_ablation_gaokao}
\end{table}

\begin{table}[t]
\centering
\small
\begin{tabular}{lcc}
\toprule
Metric & Correct verdict & Incorrect verdict \\
\midrule
Mean calls & 1.60 & 1.34 \\
Median calls & 1.0 & 0.0 \\
Success & 88.2\% & 87.3\% \\
Python Error & 9.6\% & 10.6\% \\
Quota Exceeded  & 2.1\% & 2.0\% \\
Timeout & $<0.1\%$ & $<0.1\%$ \\
\bottomrule
\end{tabular}
\caption{\textbf{Tool execution statistics.} We compare tool-usage patterns between samples with correct and incorrect final verifier verdicts.}
\label{tab:tool_stats}
\end{table}

We provide additional analysis of tool usage in Agentic Verifier. 
Table~\ref{tab:tool_ablation_math} and Table~\ref{tab:tool_ablation_gaokao} compares the Base model, Agentic Verifier without tools, and Agentic Verifier with Python tools. 
The results show that the without-tool variant already substantially improves over the Base model, indicating that the gains do not come solely from Python execution, but also from the structured verification process itself. 
Adding the Python tool improves performance on both MATH500 and Gaokao2023, showing that external grounding provides additional benefits on top of the verifier logic.

Table~\ref{tab:tool_stats} further summarizes tool execution behavior during inference. 
We cap tool usage at 3 Python calls per sample. 
A successful tool call is defined as one that finishes normally and returns an executable result, i.e., without Python/runtime errors, timeout, or being blocked by the call budget. 
Overall, tool usage remains moderate and stable. Tool usage is slightly more frequent for correct verdicts , while the execution outcome breakdown remains broadly similar across the two groups. This suggests that correct verification tends to benefit from somewhat more tool grounding.

\subsection{Experimental Details for Generalization Beyond Math}
\label{sec:beyond_math}
This section summarizes the experimental setup for the generalization analysis beyond mathematical reasoning.

\paragraph{Benchmarks.}
We use two complementary benchmarks: LiveCodeBench for code generation and HotpotQA for multi-hop question answering.

\paragraph{Tool Interfaces.}
For LiveCodeBench, verification is grounded by Python execution.
For HotpotQA, verification is grounded by web-search-based retrieval.

\paragraph{Actor and Rollouts.}
We use Qwen3-8B as the actor to sample rollouts for both benchmarks.
For HotpotQA, 150 randomly selected queries are used.

\paragraph{Evaluation Protocol.}
Different from the main TTS experiments, this setting isolates verifier capability rather than evaluating the full test-time scaling pipeline.
We report evaluation accuracy on sampled rollouts, i.e., whether the verifier correctly judges a sampled rollout.
At inference time, the verifier does not observe the ground-truth answer and only inspects the sampled rollout to produce a verdict.

\subsection{Data Contamination Check}
We additionally examine whether the training sources overlap with the evaluation benchmarks. Based on the documented provenance of the training datasets, we do not find evidence that the evaluation benchmarks are included in training. Specifically, all three data sources trace their competition data only up to AIME 2023, thus excluding AIME 2024. GSM8K and Gaokao2023 are not included in any declared source dataset. While MATH appears in Light-R1 (a component of AReaL-boba-106k), it explicitly removes MATH500 from training. Therefore, we do not find direct overlap between the reported evaluation benchmarks and the training data used in this work.

\section{Case Study}
\label{sec:appendix_case}

This section presents representative case studies to qualitatively analyze
the behavior of different verification methods.

\subsection{Generative Verifier}

\begin{exbox}{Question}
Two runners, $A$ and $B$, start at a point $O$ on a linear track and run in the same direction.
Runner $B$ runs three times as fast as runner $A$.
An observer stands at point $P$ such that $\overline{OP}$ is perpendicular to the track.
Find the maximum value of $\angle APB$, in degrees.
\end{exbox}

\begin{exbox}{Candidate Solution}
The solution assumes runner $A$ and runner $B$ move at speeds $1$ and $3$ units per second,
respectively.
Let $d$ denote the distance from point $P$ to the starting point $O$.
After $t$ seconds, runner $A$ has traveled $t$ units and runner $B$ has traveled $3t$ units.
The solution models the distances from $P$ to $A$ and $B$ as $d - t$ and $3d - 3t$,
and applies the Pythagorean theorem to derive
\[
d^2 = (d - t)^2 + (3d - 3t)^2.
\]
Solving this equation yields $d = t$, and the solution concludes that
$\angle APB = 90^\circ$.
\end{exbox}

Note that the verifier's original reasoning is lengthy and difficult to follow.
For clarity, we provide a concise summary of the core reasoning ideas,
while retaining the key computational steps used by the verifier.

\begin{exbox}{Generative Verifier}
<think>
To verify the solution, proceed step by step by examining the geometric modeling.

First, model the geometry explicitly.
Align the track with the $x$-axis and place point $P$ at $(0,d)$.
After $t$ seconds, runner $A$ and runner $B$ are located at $(t,0)$ and $(3t,0)$,
respectively.

Next, compute the distances from point $P$ to the runners.
These distances follow directly from the coordinate setup:
\[
PA = \sqrt{t^2 + d^2},
\]
\[
PB = \sqrt{(3t)^2 + d^2}.
\]

Then, compare these expressions with the distances assumed in the solution.
The solution instead uses $d - t$ and $3d - 3t$, which would only be valid if the
runners were moving directly toward point $P$.
This does not match the given motion, where both runners move along the track
away from the starting point $O$.

Next, examine the equation derived in the solution.
Using the assumed distances, the solution proposes:
\[
d^2 = (d - t)^2 + (3d - 3t)^2.
\]
With the correct distance expressions, this equation cannot be obtained from a
valid application of the Pythagorean theorem to points $A$, $B$, and $P$.

Finally, check the implication of the condition $d = t$ obtained in the solution.
Under this condition, both runners coincide at point $O$, causing points $A$ and
$B$ to overlap.
In this degenerate configuration, the angle $\angle APB$ is not defined, rather
than being $90^\circ$.
</think>

<verify>
The solution incorrectly models the distances from point $P$ to the runners.
With the track aligned to the $x$-axis and $P$ at $(0,d)$, the correct distances are
\[
PA = \sqrt{t^2 + d^2}, \qquad PB = \sqrt{(3t)^2 + d^2}.
\]
Replacing these with $d - t$ and $3d - 3t$ leads to an invalid application of the
Pythagorean theorem and produces a degenerate configuration in which the angle
$\angle APB$ is not well-defined.
Therefore, the angle maximization argument in the solution is incorrect.
</verify>

<answer>
\textbf{false}
</answer>
\end{exbox}

\subsection{Forward Verifier}

The forward verifier performs multi-turn verification by sequentially inspecting
the candidate solution and identifying potential issues at each step.
Below we present the intermediate outputs produced during this process.
Note that the original thinking traces are often lengthy.
For clarity, we summarize the core reasoning ideas while explicitly retaining
the key computations whenever they are involved.

\begin{exbox}{Question}
The equation
\[
x^{10} + (13x - 1)^{10} = 0
\]

has ten complex roots

$r_1, \overline{r}_1, r_2, \overline{r}_2, r_3, \overline{r}_3,
 r_4, \overline{r}_4, r_5, \overline{r}_5$,
where the bar denotes complex conjugation.
Find the value of
\[
\frac{1}{r_1 \overline{r}_1}
+ \frac{1}{r_2 \overline{r}_2}
+ \frac{1}{r_3 \overline{r}_3}
+ \frac{1}{r_4 \overline{r}_4}
+ \frac{1}{r_5 \overline{r}_5}.
\]
\end{exbox}
\begin{exbox}{Candidate Solution}
Note that if $r$ is a root of $x^{10}+(13x-1)^{10}=0$, then $(13r)^{-1}$ is also a root.
Hence, the roots can be paired as $\bigl(r_k,(13r_k)^{-1}\bigr)$ for $k=1,\dots,5$.

For each such pair, note that
\[
\left((13r_k)^{-1}(13\overline r_k)^{-1}\right)^{-1}
= 13^2 r_k\overline r_k .
\]
Hence the paired contribution is
\[
\frac{1}{r_k\overline r_k} + 13^2 r_k\overline r_k .
\]
Summing over all five pairs yields
\[
\sum_{i=1}^{5}\frac{1}{r_i\overline r_i}
= \sum_{k=1}^{5}\Bigl((r_k\overline r_k)^{-1} + 13^2 r_k\overline r_k\Bigr).
\]

By Vieta’s formulas, the sum of the roots is $0$, which implies that the sum of
their squares is also $0$. Consequently,
\[
\sum_{k=1}^{5} r_k\overline r_k = 0,
\]
and therefore
\[
\sum_{i=1}^{5}\frac{1}{r_i\overline r_i} = 0.
\]

The answer is $\boxed{0}$.
\end{exbox}

\begin{exbox}{Forward Verifier — Stage A: Plan}

<think>
To start Stage A, first identify what the problem asks and what the candidate
solution assumes.

The problem gives a degree-10 polynomial equation
\[
x^{10} + (13x - 1)^{10} = 0
\]
with roots occurring in complex conjugate pairs. The target is
\[
\sum_{i=1}^{5} \frac{1}{r_i \overline{r_i}},
\]
which is the sum of reciprocals of squared moduli. This quantity is typically
approached via algebraic identities about roots, rather than solving roots explicitly.

The candidate solution relies on two main structural moves:
(1) a claimed root transformation: if \(r\) is a root then \(\frac{1}{13r}\) is also a root,
and uses it to pair roots and rewrite the target sum;
(2) an application of Vieta’s formulas to assert certain symmetric sums are zero,
which then leads to a denominator claimed to be zero.

Since later steps depend on these claims, Stage B should verify them in the same
order: validate the claimed root mapping first, then inspect how the pairing is used,
then check the Vieta-based statements that produce the zero denominator.
</think>

Stage B will follow these checkable steps:

Step 1. Root transformation check:
    Test whether “if \(r\) is a root, then \(\frac{1}{13r}\) is also a root” holds by direct substitution.

Step 2. Pairing-and-rewrite dependency check:
    Trace how the solution uses the claimed mapping to pair roots and rewrite the target expression,
    and identify which later equalities depend on this pairing.

Step 3. Vieta/coefficient consistency check:
    Verify whether the solution’s Vieta-based claims (e.g., “sum of roots is 0”) match the actual polynomial coefficients.

Step 4. Zero-denominator justification check:
    Examine the step that introduces a denominator asserted to be 0, and determine whether that “0” follows from valid identities or from earlier invalid assumptions.

\textbf{Stage A complete.}

\end{exbox}

\begin{exbox}{Forward Verifier — Stage B: Validation}

\textbf{Step 1}

<think>
Start with the claimed root transformation. The solution asserts:

If \(r\) is a root of
\[
x^{10} + (13x - 1)^{10} = 0,
\]
then \(s = \frac{1}{13r}\) is also a root.

To test this claim, substitute \(s = \frac{1}{13r}\) into the polynomial.

Compute:
\[
s^{10} = \left(\frac{1}{13r}\right)^{10} = \frac{1}{13^{10} r^{10}}.
\]

Also,
\[
13s - 1 = 13 \cdot \frac{1}{13r} - 1 = \frac{1}{r} - 1 = \frac{1-r}{r},
\]
so
\[
(13s - 1)^{10} = \left(\frac{1-r}{r}\right)^{10} = \frac{(1-r)^{10}}{r^{10}}.
\]

Thus,
\[
s^{10} + (13s - 1)^{10}
= \frac{1 + 13^{10}(1-r)^{10}}{13^{10} r^{10}}.
\]

For \(s\) to be a root, this expression must equal \(0\), which requires
\[
1 + 13^{10}(1-r)^{10} = 0
\]
\[
(1-r)^{10} = -\frac{1}{13^{10}}.
\]

However, the original root condition is
\[
r^{10} + (13r - 1)^{10} = 0
\]

\[
(13r - 1)^{10} = -r^{10},
\]
which is not the same constraint as \((1-r)^{10} = -\frac{1}{13^{10}}\).
So the substitution does not show that \(s=\frac{1}{13r}\) must satisfy the original equation.
</think>

The direct substitution yields
\[
s^{10} + (13s - 1)^{10} = \frac{1 + 13^{10}(1-r)^{10}}{13^{10} r^{10}},
\]
so \(s\) is a root only if \((1-r)^{10} = -\frac{1}{13^{10}}\), which does not follow from
\(r^{10} + (13r - 1)^{10} = 0\). The claimed “\(r \mapsto \frac{1}{13r}\)” root transformation is therefore not justified by the solution’s argument.
<step/>
\end{exbox}

\begin{exbox}{Forward Verifier — Stage B: Validation}

\textbf{Step 2}

<think>
Next, inspect how the solution uses the claimed mapping. It pairs terms as
\[
\frac{1}{r\overline{r}} + \frac{1}{\frac{1}{13r}\cdot\frac{1}{13\overline{r}}}
= \frac{1}{r\overline{r}} + \frac{13^{2}}{r\overline{r}}.
\]

Algebraically, the simplification
\[
\frac{1}{\frac{1}{13r}\cdot\frac{1}{13\overline{r}}} = \frac{13^2}{r\overline{r}}
\]
is fine, but this step implicitly assumes that both \(r\) and \(\frac{1}{13r}\) are among the roots being summed over.
Therefore, even if the algebra inside the term is correct, the pairing is only valid if the root-mapping claim is correct and the roots can indeed be partitioned into such pairs.

Since the previous step did not establish the root transformation, the correctness of the pairing—and hence the rewritten form of the target sum—depends on an unverified assumption.
</think>

The rewrite
\[
\frac{1}{r\overline{r}} + \frac{1}{\frac{1}{13r}\cdot\frac{1}{13\overline{r}}}
= \frac{1}{r\overline{r}} + \frac{13^{2}}{r\overline{r}}
\]
is algebraically valid, but the use of this identity inside the target sum relies on the unverified claim that \(\frac{1}{13r}\) is also a root paired with \(r\). Thus, the subsequent reduction of the five-term sum to a smaller grouped sum is not justified without establishing the root mapping.
<step/>

\end{exbox}

\begin{exbox}{Forward Verifier — Stage B: Validation}

\textbf{Step 3}

<think>
Now inspect the Vieta-based claim. The solution states:

“By Vieta’s formulas, the sum of the roots is \(0\), so the sum of the squares of the roots is \(0\) as well.”

To check the “sum of roots is 0” statement, consider the polynomial
\[
P(x) = x^{10} + (13x - 1)^{10}.
\]
The sum of roots equals
\[
-\frac{[x^9]\;P(x)}{[x^{10}]\;P(x)}.
\]

Compute the leading coefficient:
\[
[x^{10}]\;P(x) = 1 + 13^{10}.
\]

Compute the \(x^9\) coefficient: the term \(x^{10}\) contributes \(0\) to \(x^9\).
From \((13x-1)^{10}\), the \(x^9\) term comes from choosing \(9\) factors of \(13x\) and one factor of \((-1)\):
\[
[x^9]\;(13x-1)^{10} = \binom{10}{9} 13^9 (-1)^{1}.
\]
Thus,
\[
[x^9]\;P(x) = -10\cdot 13^9.
\]

Therefore the sum of roots is
\[
-\frac{-10\cdot 13^9}{1+13^{10}} = \frac{10\cdot 13^9}{1+13^{10}},
\]
which is not \(0\).
So the solution’s Vieta statement “sum of roots is 0” does not match the polynomial coefficients.
</think>

For \(P(x)=x^{10}+(13x-1)^{10}\),
\[
[x^{10}]P(x)=1+13^{10},
\]
\[
[x^9]P(x)=-10\cdot 13^9,
\]
so the sum of roots is
\[
-\frac{[x^9]P(x)}{[x^{10}]P(x)}=\frac{10\cdot 13^9}{1+13^{10}}\neq 0.
\]
Hence the solution’s claim that the sum of the roots is \(0\) is inconsistent with Vieta’s formulas for this polynomial.
<step/>

\end{exbox}

\begin{exbox}{Forward Verifier — Stage B: Validation}

\textbf{Step 4}

<think>
Finally, examine the step where the solution forms a fraction with a denominator it asserts equals \(0\), and concludes the entire expression equals \(0\).

That step depends on two separate requirements:
(1) the earlier pairing/rewrite must be valid (which depended on the root transformation);
(2) the denominator being set to \(0\) must follow from correct identities about the roots of the polynomial.

From the previous turn, the claim “sum of roots is 0” is not supported by the actual coefficients.
So any subsequent inference that a symmetric sum vanishes (and in particular a denominator equals \(0\)) lacks justification from Vieta’s formulas as applied to this polynomial.

Moreover, even conceptually, \(r\overline{r} = |r|^2\) is nonnegative real for any complex root \(r\neq 0\),
so a sum of several such terms being exactly \(0\) would force each term to be \(0\), which would require \(r=0\).
This would need to be consistent with the original equation, and cannot be assumed without verification.
</think>

The “division by 0” step relies on unsupported claims. The polynomial coefficients do not support the statement that the sum of roots is \(0\), so the subsequent denominator being asserted as \(0\) is not justified by Vieta’s formulas. In addition, since \(r\overline{r}=|r|^2\ge 0\) for each nonzero root, forcing a sum of such terms to be \(0\) would require each \(|r|^2=0\), i.e., \(r=0\), which must be checked against the original equation rather than assumed.
\texttt{<end\_of\_analysis/>}

\end{exbox}
\begin{exbox}{Forward Verifier — Stage C: Verdict}

<review>
The candidate solution is invalid for two independent reasons.

(1) The solution assumes a root-closure property: if \(r\) is a root, then
\(\frac{1}{13r}\) is also a root. Direct substitution shows that
\(s=\frac{1}{13r}\) satisfies the equation only under an additional constraint on
\(r\), which is not implied by \(r^{10}+(13r-1)^{10}=0\). Therefore the proposed
root pairing is not justified.

(2) The solution incorrectly applies Vieta’s formulas by claiming the sum of the
roots is \(0\). For \(P(x)=x^{10}+(13x-1)^{10}\), the coefficients imply the sum of
roots is \(\frac{10\cdot 13^9}{1+13^{10}}\neq 0\). As a result, later deductions
(including the step that introduces a denominator asserted to be \(0\)) do not
follow from valid polynomial identities.

Since the key structural claims fail, the final conclusion \(0\) is not supported.
</review>

<answer>
\textbf{false}
</answer>

\end{exbox}

\subsection{Backward Verifier}

\begin{exbox}{Question}
The product of $3t^2+5t+a$ and $4t^2+bt-2$ is $12t^4+26t^3-8t^2-16t+6$. What is $a+b$?
\end{exbox}

\begin{exbox}{Candidate Solution}
We can expand the product of the two polynomials to get
Let
\[
E(t)=(3t^2+5t+a)(4t^2+bt-2).
\]
Then
\[
\begin{aligned}
E(t)
&= 12t^4 + (3b+20)t^3 \\
&\quad + (4a+5b-10)t^2 \\
&\quad + (-2a-4b+10)t - 2a.
\end{aligned}
\]

Since this is equal to $12t^4+26t^3-8t^2-16t+6$, we can equate the coefficients of corresponding powers of $t$ to get the following system of equations:
\begin{align*}
3b+20 &= 26, \\
4a+5b-10 &= -8, \\
-2a-4b+10 &= -16, \\
-2a &= 6.
\end{align*}
From the last equation, we have $a=-3$.
Substituting this into the third equation, we get $-2(-3)-4b+10=-16$, which simplifies to $-4b=-22$ or $b=\frac{11}{2}$.
Therefore, $a+b=-3+\frac{11}{2}=\boxed{\frac{5}{2}}$.

The answer is: $\frac{5}{2}$

\end{exbox}

\begin{exbox}{Backward Verifier — Stage A: Plan}

\texttt{<think>}
The candidate solution claims the final value $a+b=\frac{5}{2}$.
For this to be justified, several dependencies must hold in reverse order.

If the final value is correct, then the solution must have obtained correct
values of $a$ and $b$.
Those values must satisfy a valid coefficient-matching system.
And that system must come from a correct expansion of
$(3t^2+5t+a)(4t^2+bt-2)$.

So the backward checks are:
(1) expansion correctness;
(2) coefficient equations correctness;
(3) consistency of the reported $(a,b)$ with the verified equations.
\texttt{</think>}

\noindent
\textbf{Backward verification plan}
\begin{itemize}
  \item Step 1: Verify the polynomial expansion the solution relies on (especially $t^2$ and $t$ coefficients).
  \item Step 2: Verify the coefficient-matching system derived from that expansion matches $12t^4+26t^3-8t^2-16t+6$.
  \item Step 3: Check whether the reported $a=-3$ and $b=\frac{11}{2}$ satisfy the verified system, and whether they imply $a+b=\frac{5}{2}$.
\end{itemize}

\end{exbox}

\begin{exbox}{Backward Verifier — Stage B: Validation}

\textbf{Step 1}

\texttt{<think>}
To support the final claim, the expansion
\[
(3t^2+5t+a)(4t^2+bt-2)
\]
used by the solution must be algebraically correct.

I first re-expand the product in a structured way to see what coefficients
must appear.

Compute key contributions:
\[
3t^2\cdot(4t^2+bt-2)=12t^4+3b\,t^3-6t^2
\]
\[
5t\cdot(4t^2+bt-2)=20t^3+5b\,t^2-10t
\]
\[
a\cdot(4t^2+bt-2)=4a\,t^2+ab\,t-2a
\]

Summing like terms gives:
\[
\begin{aligned}
12t^4
&+ (3b+20)t^3 \\
&+ (4a+5b-6)t^2 \\
&+ (ab-10)t \\
&- 2a.
\end{aligned}
\]

The solution instead uses $(4a+5b-10)t^2$ and $(-2a-4b+10)t$, so I want a quick
tool check to confirm the correct symbolic expansion.
\texttt{</think>}

\texttt{<python>}

from sympy import symbols, expand

t = symbols('t')

a, b = symbols('a b')

poly = (3*t**2 + 5*t + a) * (4*t**2 + b*t - 2)

print(expand(poly))

\texttt{</python>}

\textbf{Tool Result}

[Stdout]
a*b*t + 4*a*t**2 - 2*a + 3*b*t**3 + 5*b*t**2 + 12*t**4 + 20*t**3 - 6*t**2 - 10*t

\end{exbox}

\begin{exbox}{Backward Verifier — Stage B: Validation}

\textbf{Step 2}

\texttt{<think>}
The tool output confirms the correct expansion is
\[
\begin{aligned}
12t^4
&+ (3b+20)t^3 \\
&+ (4a+5b-6)t^2 \\
&+ (ab-10)t - 2a.
\end{aligned}
\]

Now I compare this with the candidate’s claimed expansion:
\[
\begin{aligned}
12t^4
&+ (3b+20)t^3 \\
&+ (4a+5b-10)t^2 \\
&+ (-2a-4b+10)t - 2a.
\end{aligned}
\]

Two mismatches appear immediately:
\[
\text{(i) } (4a+5b-6)\neq(4a+5b-10).
\]
\[
\text{(ii) } (ab-10)\neq(-2a-4b+10).
\]

Since the coefficient expressions differ, the coefficient-matching system
written in the solution cannot be derived from the true product, so the
backward dependency from the final value is broken at the equation-derivation
stage.
\texttt{</think>}

\texttt{<step/>}

\textbf{Step 3}

\texttt{<think>}
Even if the solution’s later arithmetic were internally consistent, the
reported parameters must satisfy the \emph{verified} coefficient equations
from the correct expansion when matched to
$12t^4+26t^3-8t^2-16t+6$.

The solution reports $a=-3$ (from $-2a=6$), and $b=\frac{11}{2}$.
I check whether $b=\frac{11}{2}$ is compatible with $3b+20=26$:
\[
3\cdot\frac{11}{2}+20=\frac{33}{2}+20=\frac{73}{2}\neq 26,
\]
so the reported $b$ already fails the verified system.

Therefore the final claim $a+b=\frac{5}{2}$ is not supported by the correct
algebraic conditions.
\texttt{</think>}

\texttt{<end\_of\_analysis/>}
\end{exbox}

\begin{exbox}{Backward Verifier — Stage C: Verdict}
\texttt{<review>}
Step 1 verified the correct expansion of
$(3t^2+5t+a)(4t^2+bt-2)$ by structured term collection and a single symbolic
tool check. The correct coefficients are
$(3b+20)$ for $t^3$, $(4a+5b-6)$ for $t^2$, and $(ab-10)$ for $t$.

Step 2 showed the candidate solution uses incorrect expressions for the $t^2$
and $t$ coefficients, so the coefficient-matching system it writes does not
follow from the actual product.

Step 3 checked consistency of the reported values with the verified system
and found the claimed $b=\frac{11}{2}$ contradicts $3b+20=26$ immediately.
Thus the final claim $a+b=\frac{5}{2}$ is not justified.

Overall, the backward dependencies required for the conclusion fail due to an
incorrect expansion and inconsistent parameter values.
\texttt{</review>}

\texttt{<answer>false</answer>}
\end{exbox}

\end{document}